\documentclass[11pt]{article}

\usepackage[final]{acl}

\usepackage{times}
\usepackage{latexsym}
\usepackage{colortbl}\definecolor{mygray}{gray}{.9}

\usepackage[T1]{fontenc}

\usepackage[utf8]{inputenc}

\usepackage{microtype}

\usepackage{inconsolata}

\usepackage{graphicx}
\usepackage{subcaption}

\usepackage{amsmath}
\usepackage{pifont}
\usepackage{bbding}
\usepackage{amssymb}
\usepackage{booktabs}
\usepackage{colortbl}
\usepackage{multirow}
\usepackage{graphicx}  
\usepackage{array}     
\usepackage{makecell}
\usepackage{tabularx}  
\usepackage{adjustbox}
\usepackage{dblfloatfix}
\usepackage{enumitem}
 
\newcommand\blfootnote[1]{%
\begingroup
\renewcommand\thefootnote{}\footnote{#1}%
\addtocounter{footnote}{-1}%
\endgroup
}

\title{RRAttention: Dynamic Block Sparse Attention via Per-Head Round-Robin Shifts for Long-Context Inference}

\author{
    Siran Liu$^{1,2*}$,
    ~Guoxia Wang$^{1*}$,
    ~Sa Wang$^{1,2*}$,
    ~Jinle Zeng$^{1}$,
    ~HaoYang Xie$^{1}$,\\
    ~\textbf{Siyu Lou}$^{1}$\textbf{,}
    ~\textbf{JiaBin Yang}$^{1}$\textbf{,}
    ~\textbf{DianHai Yu}$^{1,2}$\textbf{,}
    ~\textbf{Haifeng Wang}$^{1}$\textbf{,}
    ~\textbf{Chao Yang}$^{2\dagger}$ \\ 
    \normalsize $^{1}$Baidu Inc. ~~~~
    \normalsize $^{2}$Peking University \\
    \small \texttt{\{wangguoxia, zengjinle, xiehaoyang, lousiyu, yangjiabin01, yudianhai, wanghaifeng\}@baidu.com} \\
    \small \texttt{\{liusr25, wangsa\}@stu.pku.edu.cn} ~~~~
    \small \texttt{\{chao\_yang\}@pku.edu.cn}
}

\begin{document}
\maketitle

\begin{abstract}
The quadratic complexity of attention mechanisms poses a critical bottleneck for large language models processing long contexts. While dynamic sparse attention methods offer input-adaptive efficiency, they face fundamental trade-offs: requiring preprocessing, lacking global evaluation, violating query independence, or incurring high computational overhead. We present RRAttention, a novel dynamic sparse attention method that simultaneously achieves all desirable properties through a head \underline{r}ound-\underline{r}obin (RR) sampling strategy. By rotating query sampling positions across attention heads within each stride, RRAttention maintains query independence while enabling efficient global pattern discovery with stride-level aggregation. Our method reduces complexity from $O(L^2)$ to $O(L^2/S^2)$ and employs adaptive Top-$\tau$ selection for optimal sparsity. Extensive experiments on natural language understanding (HELMET) and multimodal video comprehension (Video-MME) demonstrate that RRAttention recovers over 99\% of full attention performance while computing only half of the attention blocks, achieving 2.4$\times$ speedup at 128K context length and outperforming existing dynamic sparse attention methods.
\end{abstract}
\vspace{-5mm}

\blfootnote{$^*$Equal contribution. Work done as an intern at Baidu Inc. }
\blfootnote{$^\dagger$Corresponding author.}
\section{Introduction}

The ability to process long contexts has become increasingly critical for large language models (LLMs), enabling applications such as multi-document reasoning and code repository analysis. Recent advances~~\cite{Chen2023ExtendingCW, peng2024yarn, Liu2023ScalingLO, 10.5555/3737916.3739710,ernie2025technicalreport} have pushed context windows to 128K tokens and beyond, significantly expanding the scope of addressable problems. However, the quadratic computational complexity $O(L^2)$ of the attention mechanism~\cite{ashish2017attention} poses a fundamental bottleneck, making long-context inference prohibitively expensive for deployment.
\begin{figure}[!t]  %
  \centering
  \includegraphics[width=\columnwidth]{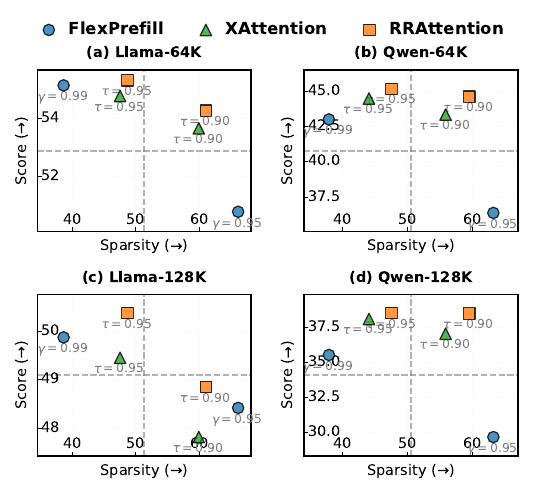}  %
  \vspace{-7mm}
  \caption{Sparsity-accuracy trade-offs across different models and context lengths on HELMET.}
  \label{fig:sdt}
  \vspace{-7mm}
\end{figure}

Sparse attention has emerged as a promising solution by selectively computing only the most important attention scores. Existing approaches can be broadly categorized into static methods that use predetermined sparsity patterns, and dynamic methods that adaptively determine sparse patterns based on input characteristics. While static methods like BigBird~\cite{zaheer2020bigbird} and StreamingLLM~\cite{xiao2023streamingllm} achieve computational efficiency, their fixed patterns cannot adapt to varying attention distributions, potentially missing critical dependencies. Dynamic methods~\cite{gao2024seerattention,jiang2024minference,lai2025flexprefill,xu2025xattention} provide input-adaptive sparsity but face different trade-offs in design choices that affect their practical effectiveness.

Through systematic analysis of existing dynamic sparse attention methods, we identify five critical dimensions that characterize their effectiveness: (i) preprocessing requirements that affect deployment flexibility, (ii) global evaluation capability for capturing long-range dependencies, (iii) query independence for maintaining semantic integrity, (iv) pattern flexibility across different attention heads, and (v) softmax granularity that impacts computational efficiency. Current methods excel in some dimensions while compromising others—no existing approach simultaneously achieves all desirable properties, leading to suboptimal trade-offs between accuracy and efficiency.

We present RRAttention, a novel dynamic sparse attention method that achieves an optimal balance across all critical dimensions through a head round-robin sampling strategy. Our key insight is that by rotating query sampling positions across attention heads within each stride, we can maintain query independence while enabling efficient global pattern discovery. Specifically, RRAttention: (i) samples one representative query per stride using a round-robin scheme across heads; (ii) performs stride-level aggregation that reduces computational cost from $O(L^2)$ to $O(L^2/S^2)$; and (iii) employs adaptive Top-$\tau$ block selection based on attention concentration patterns.

Extensive experiments demonstrate RRAttention's effectiveness across comprehensive benchmarks in natural language understanding and multimodal video comprehension. Evaluating on HELMET and Video-MME, RRAttention recovers over 99\% of full attention performance while computing only half of the attention blocks, significantly outperforming both FlexPrefill and XAttention across various sparsity settings. Furthermore, RRAttention achieves 2.4$\times$ end-to-end speedup at 128K context length and reduces pattern search overhead by 18.2\% compared to XAttention, making it both the fastest and most accurate sparse attention method in our evaluation. The main contributions of this paper are:
\begin{itemize}[itemsep=0.02em]
\item We provide a systematic characterization of dynamic sparse attention methods across five critical dimensions, revealing fundamental trade-offs in existing approaches.
\item We introduce RRAttention, a novel head round-robin sampling strategy for the attention module that uniquely achieves all desirable properties: preprocessing-free deployment, global evaluation, query independence, pattern-agnostic simplicity, and efficient stride-level computation.
\item Comprehensive experiments on both language (HELMET) and multimodal (Video-MME) benchmarks validate our approach's superiority in balancing efficiency and accuracy across diverse long-context scenarios.
\end{itemize}

\section{Preliminary and Background}
\subsection{Block-wise Sparse Attention}

For an input sequence of length $L$, the query matrix $\mathbf{Q} \in \mathbb{R}^{L \times d}$ and key matrix $\mathbf{K} \in \mathbb{R}^{L \times d}$ are partitioned into consecutive blocks of size $B$, where $d$ denotes the attention head dimension. Distinctly, we define a finer sampling stride $S$ ($S \ll B$) for pattern discovery. Let $N_b = \lceil L/B \rceil$ be the total number of blocks. The $i$-th query block and $j$-th key block are denoted as $\mathbf{Q}^{(i)}$ and $\mathbf{K}^{(j)}$ respectively, where $i, j \in \{1, 2, \ldots, N_b\}$.

In block-wise sparse attention~\citep{guo2024blocksparse}, a block selection matrix $\mathbf{B} \in \{0,1\}^{N_b \times N_b}$ is defined, where $\mathbf{B}_{ij} = 1$ indicates that attention scores are computed between the $i$-th query block and the $j$-th key block, while $\mathbf{B}_{ij} = 0$ indicates that the computation for this block pair is skipped. The sparse attention computation is formalized as:
\begin{equation}
    \mathbf{A}_{\text{sparse}} = \text{Softmax}\left(\frac{1}{\sqrt{d}}\mathbf{Q}\mathbf{K}^T \odot \mathbf{M}_\mathbf{B}\right)\mathbf{V}
\end{equation}

where $\mathbf{M}_\mathbf{B} \in \mathbb{R}^{L \times L}$ is the fine-grained mask matrix expanded from the block selection matrix $\mathbf{B}$:
\begin{equation}
    \mathbf{M}_\mathbf{B}[i,j] = \begin{cases}
        1, & \text{if } \mathbf{B}_{\lfloor i/B \rfloor, \lfloor j/B \rfloor} = 1 \text{ and } j \leq i \\
        0, & \text{otherwise}
    \end{cases}
\end{equation}

The optimization objective seeks to minimize the approximation error between sparse and full attention outputs while constraining computational cost:
\begin{align}
    \min_{\mathbf{B}} \quad & \|\mathbf{A}_{\text{full}} - \mathbf{A}_{\text{sparse}}\|_F \\
    \text{s.t.} \quad & \sum_{i,j} \mathbf{B}_{ij} \leq (1- \rho) \cdot N_b^2 \\
    & \mathbf{B}_{ij} = 0, \quad \forall j > i
\end{align}

where $\|\cdot\|_F$ denotes the Frobenius norm, $\rho \in (0,1]$ is a sparsity parameter, and the second constraint ensures causality for autoregressive models.

\subsection{Multi-Dimensional Analysis of Attention Selection}
\begin{table*}[!htbp]
\centering
\caption{Comparison of dynamic sparse attention methods}
\label{tab:dynamic_methods}
\begin{tabular}{lccccc}
\toprule
Method & \begin{tabular}[c]{@{}c@{}}Preprocessing\\-free\end{tabular} & \begin{tabular}[c]{@{}c@{}}Global\\Evaluation\end{tabular} & \begin{tabular}[c]{@{}c@{}}Query\\Independence\end{tabular} & \begin{tabular}[c]{@{}c@{}}Pattern\\-agnostic\end{tabular} & \begin{tabular}[c]{@{}c@{}}Softmax\\Granularity\end{tabular} \\
\midrule
SeerAttention & \ding{55} & \checkmark & \checkmark & \checkmark & Block \\
MInference & \ding{55} & \ding{55} & \checkmark & \ding{55} & Token \\
FlexPrefill & \checkmark & \ding{55} & \checkmark & \ding{55} & Token \\
XAttention & \checkmark & \checkmark & \ding{55} & \checkmark & Stride \\
\midrule
\textbf{RRAttention} & \checkmark & \checkmark & \checkmark & \checkmark & Stride \\
\bottomrule
\end{tabular}
\end{table*}

Sparse attention mechanisms can be broadly categorized into static and dynamic patterns. Static sparse methods, such as BigBird~\cite{zaheer2020bigbird}, StreamingLLM~\cite{xiao2023streamingllm}, and TriangleMix~\cite{he2025trianglemixlosslessefficientattention}, employ a predetermined block selection matrix $\mathbf{B}_{\text{static}}$ that remains unchanged regardless of the input sequence. This design effectively reduces computational complexity. However, recent studies have shown that the optimal sparsity pattern can vary significantly across different inputs and layers. Because static patterns cannot adapt $\mathbf{B}$ based on input-specific characteristics of $\mathbf{Q}$ and $\mathbf{K}$, they are fundamentally limited in their ability to capture critical or long-range dependencies.

In contrast, dynamic sparse attention methods adaptively determine the block selection matrix $\mathbf{B}_{dynamic}$ according to the current input, allowing the attention pattern to flexibly adjust and better suit the data at hand. By taking into account input-specific features, dynamic methods can potentially achieve higher accuracy and capture more relevant dependencies than static approaches. Table~\ref{tab:dynamic_methods} summarizes the key characteristics of existing dynamic sparse attention methods.

\textbf{SeerAttention}~\cite{gao2024seerattention} employs block-wise pooling of $\mathbf{Q}$ and $\mathbf{K}$ matrices followed by learned linear transformations through an AttnGate module trained via distillation. \textbf{MInference}~\cite{jiang2024minference} offline assigns optimal sparse patterns to each attention head, then dynamically constructs sparse indices during inference using the last query block for vertical-slash heads or mean pooling for block-sparse heads. \textbf{FlexPrefill}~\cite{lai2025flexprefill} also leverages the last query block for pattern discovery but adaptively determines query-aware or vertical-slash patterns using Jensen-Shannon divergence as a reliability metric. \textbf{XAttention}~\cite{xu2025xattention} samples anti-diagonal elements at stride granularity and achieves summation through dimension aggregation to estimate importance across the entire attention matrix.
These methods exhibit distinct trade-offs across the following dimensions:

\begin{itemize}
\item \textit{Preprocessing-free} indicates whether a method requires offline training or pattern search. Methods requiring preprocessing (SeerAttention's distillation training, MInference's offline pattern assignment) achieve better accuracy but limit deployment flexibility, while preprocessing-free methods (FlexPrefill, XAttention) offer immediate applicability to new models and tasks.

\item \textit{Global evaluation} determines whether importance estimation considers the entire query-key space. Global methods (SeerAttention, XAttention) capture long-range dependencies more effectively but incur higher computational overhead, whereas local methods (MInference, FlexPrefill) that only use the last query block are more efficient but may miss important patterns in earlier positions.

\item \textit{Query independence} ensures each query's attention distribution is computed separately without cross-query interference. Methods violating this principle (XAttention's anti-diagonal aggregation across different queries) risk mixing attention distributions from different contexts, potentially introducing evaluation errors, while query-independent methods preserve the semantic integrity of each query's attention pattern.

\item \textit{Pattern-agnostic} reflects whether a method applies a unified strategy across all attention heads. Pattern-specific methods (MInference, FlexPrefill) achieve better accuracy by tailoring to different attention patterns but require additional pattern detection overhead, while pattern-agnostic approaches (SeerAttention, XAttention) offer simpler implementation and consistent behavior.

\item \textit{Softmax granularity} significantly impacts computational efficiency: block-level or stride-level aggregation (SeerAttention, XAttention) accelerates pattern discovery through dimension reduction but may lose fine-grained information, while token-level softmax (MInference, FlexPrefill) preserves precise attention distributions at the cost of higher computational complexity.
\end{itemize}

RRAttention presents a balanced solution that combines the strengths of existing approaches: it requires no preprocessing for maximum flexibility, performs global evaluation for comprehensive pattern capture, maintains query independence for accurate attention computation, applies a pattern-agnostic strategy for simplicity, and uses stride-level softmax for computational efficiency. This design achieves an optimal trade-off between accuracy and efficiency for dynamic sparse attention in long-context scenarios.

\begin{figure*}[!htbp]
  \centering
  \includegraphics[width=1\linewidth]{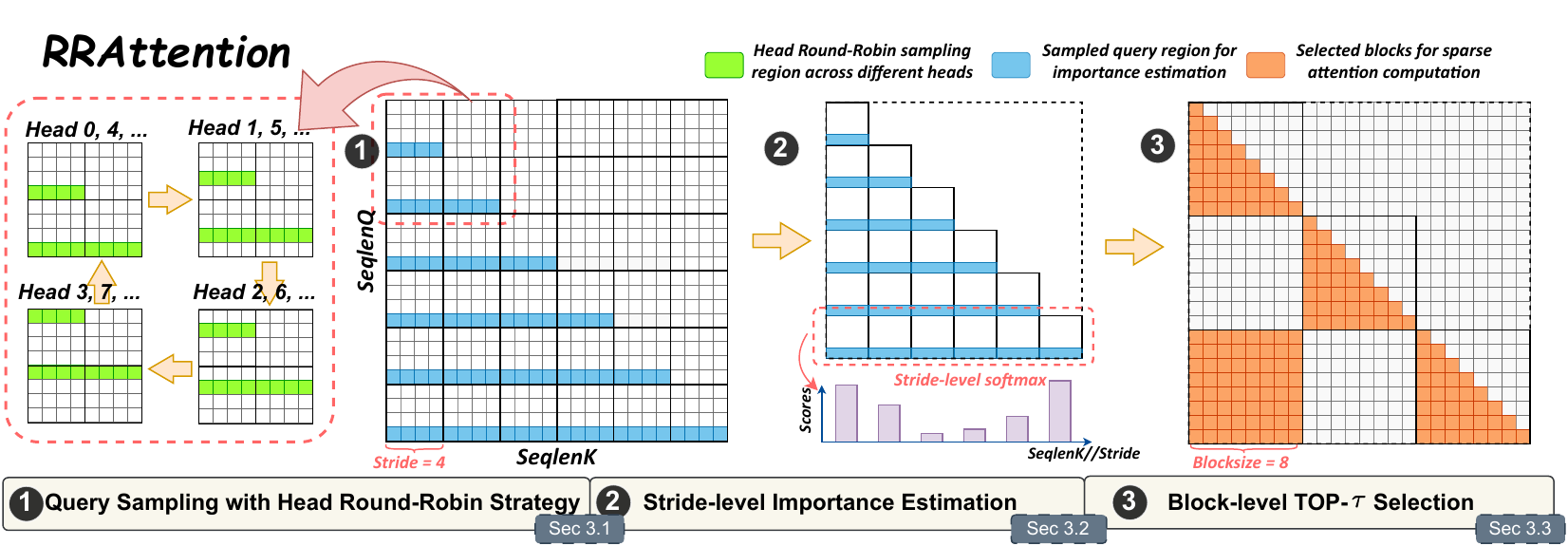}
  \caption{Illustration of RRAttention. \ding{172}, \ding{173}, and \ding{174} represent the three stages of our method. The example shows a configuration with stride size $S=4$ and block size $B=8$.}
  \label{fig:method}
  \vspace{-5mm}
\end{figure*}

\section{Methodology: RRAttention}
\label{sec:Methodology}

RRAttention achieves efficient dynamic sparse attention through a novel head round-robin sampling strategy that maintains query independence while enabling fast pattern discovery. Figure~\ref{fig:method} illustrates the complete workflow of RRAttention.

\subsection{Query Sampling with Head Round-Robin Strategy}
To maintain query independence, RRAttention samples one representative query per stride $S$ tokens to compute attention with all keys, ensuring that each query's attention distribution is computed separately without cross-query interference. This preserves the semantic integrity of each query's attention pattern. For attention head $h$, the sampled query position $P(i, h)$ within each stride interval $[iS, (i+1)S)$ is determined by:
\begin{equation}
    P(i, h) = iS + (S - 1 - (h \bmod S))
\end{equation}

The head round-robin strategy serves two critical purposes: (i) it ensures that across different heads, all positions within a stride are eventually sampled, preventing information loss that would occur with fixed-position sampling; (ii) the varied sampling positions naturally intersect with important vertical and slash patterns in the attention matrix, enabling effective detection of these critical structures. The sampled query set for head $h$ is:
{\fontsize{10pt}{14pt}\selectfont
\begin{equation}
\begin{aligned}
\mathbf{Q}^{(h)}_{\text{sample}} = \{&\mathbf{Q}_{j} : j = P(i, h), i \in [0, \lfloor L/S \rfloor)\}
\end{aligned}
\end{equation}
}
\subsection{Stride-level Importance Estimation}

To accelerate pattern discovery, we aggregate attention scores at stride granularity through dimension reduction. For query stride $i$ and key stride $j$, we compute an importance score representing their aggregated attention by first aggregating key vectors within each stride and then computing the dot product with the sampled query:
\begin{equation}
    \mathbf{I}_{i,j}^{(h)} = \frac{1}{S\sqrt{d}} \mathbf{Q}_{P(i,h)}^{(h)} \cdot \left(\sum_{k=jS}^{(j+1)S-1} \mathbf{K}_k^{(h)}\right)
\end{equation}

where $\mathbf{Q}_{P(i,h)}$ is the query vector at the sampled position from stride $i$ for head $h$, and the summation aggregates all key vectors within stride $j$. The normalization factor $\frac{1}{S\sqrt{d}}$ accounts for both attention scaling and stride size.

We then apply row-wise softmax to obtain normalized importance scores for each query stride:
\begin{equation}
    \mathbf{P}_{i,j}^{(h)} = \frac{\exp(\mathbf{I}_{i,j}^{(h)})}{\sum_{j'=0}^{\lfloor L/S \rfloor-1} \exp(\mathbf{I}_{i,j'}^{(h)})}
\end{equation}

Through query sampling and key aggregation, this stride-level importance estimation reduces the computational complexity from $O(L^2)$ to $O(L^2/S^2)$, while maintaining global evaluation capability as every position in the sequence contributes through our head-robin sampling strategy.

\subsection{Block-level Selection via Top-$\tau$ Thresholding}

To align with our block-wise sparse attention formulation and enable efficient GPU computation, we aggregate stride-level importance scores to block level. For query block $m$ and key block $n$, the block importance score is computed as:
\begin{equation}
    \mathbf{S}_{m,n}^{(h)} = \sum_{\substack{i \in \text{block } m \\ j \in \text{block } n}} \mathbf{P}_{i,j}^{(h)}
\end{equation}

For each query block $m$, we employ a top-$\tau$ selection strategy that retains blocks whose cumulative importance exceeds threshold $\tau$. Specifically, we sort key blocks by importance and select the minimal set such that:
{\fontsize{9pt}{14pt}\selectfont
\begin{equation}
    \mathbf{B}^{\text{dynamic}}_{m,n} = \begin{cases}
        1, & \text{if } n \in \{k : \sum_{l \in \text{sort}(\mathbf{S}_{m,*}^{(h)})[1:k]} \mathbf{S}_{m,l}^{(h)} \geq \tau\} \\
        0, & \text{otherwise}
    \end{cases}
\end{equation}
}

Following prior work demonstrating the critical importance of final query positions for maintaining generation quality, we construct a protected mask $\mathbf{B}^{\text{static}}$ that unconditionally retains all blocks for the last query block. The final block selection matrix combines both components:
\begin{equation}
    \mathbf{B}_{m,n} = \mathbf{B}^{\text{dynamic}}_{m,n} \lor \mathbf{B}^{\text{static}}_{m,n}
\end{equation}

{\section{Experiments}
\label{sec:Experiments}

\subsection{Settings}
\textbf{Models.} For natural language tasks, we evaluate two prominent long-context language models: (i) \textbf{Meta-LLaMA-3.1-8B-Instruct}~\cite{grattafiori2024llama3herdmodels}, which supports up to 128K tokens context length with strong instruction-following capabilities; (ii) \textbf{Qwen2.5-7B-Instruct}~\cite{qwen2025qwen25technicalreport}, which supports up to 128K tokens context length through YARN~\cite{peng2023yarnefficientcontextwindow} scaling with enhanced multilingual performance.
For video understanding tasks, we evaluate \textbf{Qwen2-VL-7B-Instruct}~\cite{wang2024qwen2vlenhancingvisionlanguagemodels}, a vision-language model capable of processing and understanding video content through multimodal attention. We maintain the default chat template settings for all test datasets to ensure consistent evaluation across different models.
Additionally, we provide generalization experiments on larger-scale and diverse architectures (e.g., Yi-9B-200K, Qwen3-30B-A3B) in Appendix~\ref{sec:appendix_additional_models}.

\textbf{Baselines.} We evaluate several state-of-the-art baseline methods: (i) \textbf{FlashAttention}~\cite{dao2022flashattention, dao2023flashattention2}, a highly optimized attention kernel that serves as our dense baseline for measuring the performance of standard full attention computation. (ii) \textbf{FlexPrefill}, a dynamic sparse attention method that captures the Vertical-Slash pattern through the last query block; we test it with hyperparameter $\tau=0.1$ and two threshold settings $\gamma=0.95$, and $\gamma=0.99$. (iii) \textbf{XAttention}, a dynamic method that identifies important block patterns through anti-diagonal evaluation; we evaluate it with block size $S=8$ and two threshold settings $\tau=0.9$ and $\tau=0.95$. (iv) \textbf{RRAttention}, our proposed method that employs head round-robin sampling for global pattern discovery; we evaluate it with the same configurations as XAttention.
All sparse attention experiments focus on the prefill stage, while the decoding stage employs uniform dense computation for all methods.

\textbf{Datasets.} We evaluate on the following benchmarks across two task categories:
(i) For natural language understanding tasks, we employ \textbf{HELMET}~\cite{yen2025helmet}, a comprehensive long-context benchmark that includes seven task categories: retrieval-augmented generation (Natural Questions, TriviaQA, HotpotQA), generation with citations (ALCE), passage re-ranking (MS MARCO), many-shot in-context learning (TREC, BANKING77, CLINC150), long-document QA (NarrativeQA, $\infty$Bench), summarization (Multi-LexSum), and synthetic recall (RULER, JSON KV). 
(ii) For video understanding tasks, we use \textbf{Video-MME}~\cite{fu2024video}, the first full-spectrum multimodal evaluation benchmark for MLLMs, containing 900 videos across three duration categories with 2700 multiple-choice QA pairs. The dataset requires models to comprehend temporal video content across 12 task types, including perception, reasoning, and information synthesis. 
All experiments are conducted on NVIDIA H100 GPUs.

\subsection{Main Result}
\begin{table*}[!t]
  \centering
  \caption{Performance comparison on HELMET benchmark across different context lengths.}
  \vskip -0.1in
  \label{tab:main_results}
  \resizebox{0.9\textwidth}{!}{
  \begin{tabular}{c | l | c | c c c c c | c}
    \toprule
    \textbf{Model} & \textbf{Methods} & \textbf{Avg. Sparsity(↑)} & \textbf{8K} & \textbf{16K} & \textbf{32K} & \textbf{64K} & \textbf{128K} & \textbf{Avg.(↑)} \\
    \midrule
    \multirow{7}{*}{Llama}
      & FullAttention & - (0\%) & 61.17 & 58.67 & 56.71 & 55.81 & 49.74 & 56.42 \\
    \cline{2-9}
      & \multirow{2}{*}{FlexPrefill}
                       & 0.95 (66.07\%) & 56.02 & 54.48 & 53.01 & 50.77 & 48.42 & 52.54 \\
      &                & 0.99 (38.63\%) & 60.73 & 58.15 & 56.22 & 55.15 & 49.87 & 56.02 \\
    \cline{2-9}
      & \multirow{2}{*}{XAttention}
                       & 0.90 (59.88\%) & 59.92 & 57.15 & 55.54 & 53.66 & 47.82 & 54.82 \\
      &                & 0.95 (47.50\%) & 60.79 & 58.10 & 55.57 & 54.78 & 49.45 & 55.74 \\
    \cline{2-9}
      & \multirow{2}{*}{\textbf{RRAttention}}
                       & 0.90 (61.02\%) & \textbf{60.95} & \textbf{58.43} & \textbf{55.86} & \textbf{54.27} & \textbf{48.85} & \textbf{55.67} \\
      &                & 0.95 (48.68\%) & \textbf{61.22} & \textbf{58.21} & \textbf{56.07} & \textbf{55.33} & \textbf{50.37} & \textbf{56.24} \\
    \midrule
    \multirow{7}{*}{Qwen}
      & FullAttention & - (0\%) & 54.75 & 50.66 & 48.40 & 44.98 & 39.59 & 47.68 \\
    \cline{2-9}
      & \multirow{2}{*}{FlexPrefill}
                       & 0.95 (63.21\%) & 44.11 & 40.59 & 40.60 & 36.39 & 29.66 & 38.27 \\
      &                & 0.99 (37.88\%) & 53.18 & 49.05 & 46.48 & 43.01 & 35.52 & 45.45 \\
    \cline{2-9}
      & \multirow{2}{*}{XAttention}
                       & 0.90 (59.15\%) & 52.53 & 47.72 & 45.72 & 43.35 & 37.04 & 45.27 \\
      &                & 0.95 (47.35\%) & 53.90 & 49.77 & 46.96 & 44.49 & 38.09 & 46.64 \\
    \cline{2-9}
      & \multirow{2}{*}{\textbf{RRAttention}}
                       & 0.90 (59.53\%) & \textbf{53.37} & \textbf{49.31} & \textbf{47.35} & \textbf{44.62} & \textbf{38.48} & \textbf{46.63} \\
      &                & 0.95 (47.56\%) & \textbf{54.22} & \textbf{50.02} & \textbf{47.92} & \textbf{45.17} & \textbf{38.51} & \textbf{47.17} \\
    \bottomrule
  \end{tabular}}
\end{table*}

\begin{table*}[!t]
  \centering
  \caption{Performance comparison on Video-MME benchmark using Qwen2-VL-7B-Instruct.}
  \vskip -0.1in
  \label{tab:attention_performance}
  \resizebox{0.76\linewidth}{!}{
  \begin{tabular}{c|c|c|c|c|c|c}
    \toprule
    \textbf{FPS} & \textbf{Method} & \textbf{Avg. Sparsity(↑)} & \textbf{Short} & \textbf{Medium} & \textbf{Long} & \textbf{Avg.(↑)} \\
    \midrule
    \multirow{7}{*}{\centering \textbf{1fps}}
      & FullAttention & - (0\%) & 72.90 & 63.60 & 55.20 & 63.90 \\ 
      \cline{2-7}
      & \multirow{2}{*}{\centering FlexPrefill}
                       & 0.95 (46.70\%) & \textbf{72.80} & 62.00 & 54.00 & 62.90 \\
      &                & 0.99 (17.80\%) & 72.40 & 63.60 & 55.80 & 63.90 \\
    \cline{2-7}
      & \multirow{2}{*}{\centering XAttention}
                       & 0.90 (49.30\%) & 71.80 & 62.80 & 55.70 & 63.40 \\
      &                & 0.95 (37.50\%) & \textbf{72.70} & 63.60 & 56.10 & 64.10 \\
    \cline{2-7}
      & \multirow{2}{*}{\centering \textbf{RRAttention}}
                       & 0.90 (47.00\%) & 72.20 & \textbf{63.60} & \textbf{56.20} & \textbf{64.00} \\
      &                & 0.95 (34.70\%) & 72.60 & \textbf{64.00} & \textbf{56.20} & \textbf{64.30} \\
    \midrule
    \multirow{7}{*}{\centering \makecell{\textbf{0.5fps}\\\textbf{wsubs}}}
    & FullAttention & - (0\%) & 72.10 & 68.80 & 62.40 & 67.80 \\ 
    \cline{2-7}
      & \multirow{2}{*}{\centering FlexPrefill}
                       & 0.95 (58.60\%) & 70.80 & 67.00 & 59.40 & 65.70 \\
      &                & 0.99 (28.80\%) & 71.70 & 68.40 & 61.90 & 67.30 \\
    \cline{2-7}
      & \multirow{2}{*}{\centering XAttention}
                       & 0.90 (51.50\%) & \textbf{72.00} & 68.00 & 61.10 & 67.00 \\
      &                & 0.95 (39.90\%) & \textbf{72.30} & \textbf{68.90} & \textbf{62.30} & \textbf{67.90} \\
    \cline{2-7}
      & \multirow{2}{*}{\centering \textbf{RRAttention}}
                       & 0.90 (49.30\%) & 71.70 & \textbf{68.70} & \textbf{62.20} & \textbf{67.50} \\
      &                & 0.95 (37.40\%) & 72.20 & 68.70 & \textbf{62.30} & 67.70 \\
    \bottomrule
  \end{tabular}}
  \vskip -0.1in
\end{table*}

\textbf{HELMET.} Table~\ref{tab:main_results} and Figure~\ref{fig:sdt} present the average performance across all HELMET tasks at context lengths from 8K to 128K tokens. RRAttention consistently achieves the highest accuracy among all sparse attention methods on both models. At the conservative setting ($\tau=0.95$ and $\gamma=0.99$), RRAttention achieves 56.24 on Llama, compared to XAttention's 55.74 and FlexPrefill's 56.02 (48.68\% vs 47.50\% and 38.63\% sparsity). On Qwen with the same settings, RRAttention achieves 47.17, outperforming XAttention's 46.64 and FlexPrefill's 45.45 (47.56\% vs 47.35\% and 37.88\% sparsity). Notably, RRAttention recovers 99.7\% and 99.0\% of FullAttention performance on Llama and Qwen respectively, while computing approximately half of the attention blocks.

The accuracy-sparsity trade-off advantage of RRAttention is even more pronounced at the aggressive setting ($\tau=0.90$ and $\gamma=0.95$). On Llama, RRAttention maintains 55.67, significantly outperforming XAttention's 54.82 and FlexPrefill's 52.54 (61.02\% vs 59.88\% and 66.07\% sparsity). The pattern holds on Qwen, where RRAttention achieves 46.63, compared to XAttention's 45.27 and FlexPrefill's 38.27 (59.53\% vs 59.15\% and 63.21\% sparsity). This demonstrates that RRAttention more effectively identifies critical attention patterns—achieving higher sparsity than XAttention while maintaining better accuracy, and preserving significantly better accuracy than FlexPrefill despite its higher sparsity. Detailed analysis of performance on all individual subtasks in HELMET is provided in Appendix~\ref{tab:app_helmet}.

\textbf{Video-MME.} Table~\ref{tab:attention_performance} presents results on the Video-MME benchmark. At the standard 1fps setting, RRAttention consistently achieves the highest overall accuracy under both conservative and aggressive configurations, outperforming XAttention and FlexPrefill while maintaining comparable or better sparsity levels. This advantage becomes particularly pronounced on Medium and Long videos. Unlike text-based tasks where attention patterns often concentrate on specific tokens, video understanding requires modeling complex spatiotemporal relationships distributed across frames, making RRAttention's global evaluation capability especially valuable for capturing temporal dependencies in extended sequences. At the lower 0.5fps sampling rate with subtitles, RRAttention maintains competitive performance across settings. These results validate our method's effectiveness in multimodal scenarios with diverse attention characteristics.

\textbf{Runtime Efficiency.} Figure~\ref{fig:efficiency} presents the runtime comparison on Meta-Llama-3.1-8B-Instruct across context lengths from 8K to 128K tokens. At 128K context length, RRAttention achieves a 2.4$\times$ speedup over Full Attention while delivering the fastest inference speed among all sparse attention methods. This efficiency advantage is particularly valuable given that RRAttention simultaneously achieves state-of-the-art accuracy on HELMET benchmark, as demonstrated in Table~\ref{tab:main_results}.

The superior efficiency is further evidenced by pattern search overhead. As shown in Figure~\ref{fig:efficiency}(b), our head round-robin sampling strategy combined with stride-level aggregation reduces pattern discovery time by 18.2\% compared to XAttention at 128K tokens. This improvement stems from our design that avoids redundant computation through systematic query sampling and efficient importance estimation, demonstrating RRAttention's effectiveness for practical deployment in long-context scenarios. A detailed stage-wise latency breakdown is provided in Appendix~\ref{sec:appendix_time_breakdown}.

\begin{figure*}[htbp]
  \centering
  \includegraphics[width=1\linewidth]{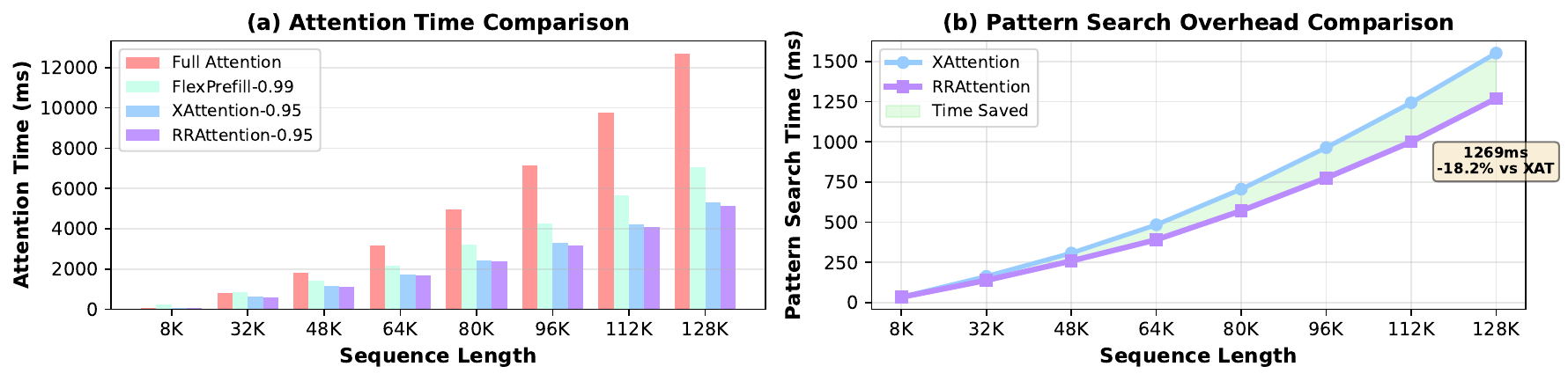}
  \vskip -0.1in
  \caption{Runtime comparison of attention methods on LLaMA-3.1-8B-Instruct across different context lengths. (a): Attention overhead. (b): Pattern search time.}
  \label{fig:efficiency}
  \vskip -0.05in
\end{figure*}

\subsection{Ablation Study}

\begin{table*}[!t]
  \centering
  \caption{Performance comparison of different strides on HELMET using LLaMA-3.1-8B-Instruct.}
  \label{tab:abl_stride}
  \resizebox{0.85\linewidth}{!}{
  \begin{tabular}{l|cc|cc|cc|cc}
    \toprule
    \multirow{2}{*}{\textbf{Method}} & \multicolumn{2}{c|}{\textbf{S=4}} & \multicolumn{2}{c|}{\textbf{S=8}} & \multicolumn{2}{c|}{\textbf{S=16}} & \multicolumn{2}{c}{\textbf{S=32}} \\
    \cmidrule(lr){2-3} \cmidrule(lr){4-5} \cmidrule(lr){6-7} \cmidrule(lr){8-9}
     & Score & Sparsity & Score & Sparsity & Score & Sparsity & Score & Sparsity \\
    \midrule
    XAttention-0.90  & 55.24 & 60.50\% & 54.82 & 60.08\% & 54.56 & 59.43\% & 54.50 & 57.69\% \\
    RRAttention-0.90  & \textbf{55.62} & \textbf{61.38\%} & \textbf{55.67} & \textbf{61.02\%} & \textbf{55.37} & \textbf{60.70\%} & \textbf{54.97} & \textbf{60.02\%} \\
    \midrule
    XAttention-0.95  & \textbf{55.89} & 48.23\% & 55.74 & 47.50\% & 55.45 & 46.46\% & 55.49 & 45.27\% \\
    RRAttention-0.95  & 55.85 & \textbf{49.03\%} & \textbf{56.24} & \textbf{48.68\%} & \textbf{55.91} & \textbf{47.96\%} & \textbf{55.62} & \textbf{47.78\%} \\
    \bottomrule
  \end{tabular}}
\end{table*}

\textbf{Last Query Block Protection.} To validate the effectiveness of protecting the last query block, we apply this strategy to XAttention baseline. As shown in Table~\ref{tab:abl_last_q_block}, adding last query block protection to XAttention yields modest improvements (from 55.74 to 55.92) with sparsity cost, confirming its utility for maintaining generation quality. However, RRAttention achieves substantially better performance (56.24), 
demonstrating that effective pattern discovery through our head round-robin strategy contributes more significantly to overall performance than the protection mechanism alone.

\textbf{Round-Robin Strategy.} To isolate the impact of round-robin(RR) scheduling, we compare performance without RR mechanism and across three RR strategies: head-RR, layer-RR, and hybrid-RR. As shown in Table~\ref{tab:abl_robin_strategy}, while all three strategies achieve comparable performance, head-RR consistently outperforms alternatives across most context lengths, demonstrating superior effectiveness in capturing critical attention patterns. These results validate that diversifying query positions across attention heads provides the most effective approach for comprehensive pattern coverage, combining both simplicity and robust performance.
\begin{table*}[!t]
  \centering \large
  \caption{Ablation study on the impact of protecting the last query block on HELMET use LLaMA-3.1-8B-Instruct.}
  \label{tab:abl_last_q_block}
  \resizebox{1\linewidth}{!}{
  \begin{tabular}{l | c | c c | c c c c c | c}
    \toprule
    \textbf{Method} &\textbf{Sparsity} & Sink \& Recent Block & Last Q Block & \textbf{8K} & \textbf{16K} & \textbf{32K} & \textbf{64K} & \textbf{128K} & \textbf{Avg.} \\
    \midrule
    XAttention-0.95 & 47.50\%  & \checkmark  &  \ding{55} & 60.79 & 58.10 & 55.57 & 54.78 & 49.45 & 55.74 \\
    XAttention+$\mathbf{B}^{\text{static}}$ & 46.40\%  & \checkmark & \checkmark & 61.09 & \textbf{58.27} & 55.94 & 54.51 & 49.77 & 55.92 \\
    RRAttention-0.95 & 49.06\% & \checkmark  & \checkmark & \textbf{61.44} & 58.19 & 55.74 & \textbf{55.51} & 50.00 & 56.18\\
    \rowcolor{mygray}RRAttention-0.95 & 48.68\%  & \ding{55}  & \checkmark & 61.22 & 58.21 & \textbf{56.07} & 55.33 & \textbf{50.37} & \textbf{56.24} \\
    \bottomrule
  \end{tabular}
  }
  \vskip -0.15in
\end{table*}

\begin{table}[!t]
  \centering
  \caption{Isolated ablation study on RR scheduling strategies on HELMET use LLaMA-3.1-8B-Instruct.}
  \label{tab:abl_robin_strategy}
  \resizebox{\columnwidth}{!}{
  \begin{tabular}{l | c c c c c c}
    \toprule
    \textbf{Method} & \textbf{8K} & \textbf{16K} & \textbf{32K} & \textbf{64K} & \textbf{128K} & \textbf{Avg.} \\
    \midrule
    w/o RR & 60.91 & 57.86 & 55.67 & 54.62 & 49.18 & 55.65 \\
    Head-RR & \textbf{60.93} & 57.90 & \textbf{56.03} & 54.70 & \textbf{49.44} & \textbf{55.80} \\
    Layer-RR & 60.75 & 57.60 & 55.99 & \textbf{54.95} & 48.40 & 55.54\\
    Hybrid-RR & 60.42 & \textbf{57.96} & 56.03 & 54.54 & 49.10 & 55.61 \\
    \bottomrule
  \end{tabular}}
  \vskip -0.15in
\end{table}

\textbf{Stride Sizes.} The stride size $S$ determines the granularity of query sampling in RRAttention, directly influencing the balance between pattern estimation accuracy and computational efficiency. As shown in Table~\ref{tab:abl_stride}, for stride configurations $S < 32$, RRAttention maintains stable performance with minimal accuracy variation and consistently high sparsity. However, at larger stride $S=32$, coarse-grained aggregation dilutes important signals from critical tokens, leading to accuracy drops. These results demonstrate that the head round-robin sampling strategy effectively captures critical attention patterns across moderate stride ranges without requiring excessively fine-grained query sampling, thereby achieving both robust pattern estimation and computational efficiency. Compared to XAttention baseline, RRAttention consistently achieves superior or comparable accuracy with higher sparsity across all stride settings.

\textbf{Block Selection Accuracy.} We validate RRAttention's importance estimation through post-hoc comparison with full attention ground truth. As detailed in Appendix~\ref{appendix:block_selection}, RRAttention achieves higher precision and F1 scores, confirming more accurate important block identification.

\section{Related Work}
\label{sec:RelatedWork}

\subsection{Training-based Sparse Methods}

Several approaches integrate sparsity into the training process. NSA~\cite{yuan2025nativesparseattentionhardwarealigned} learns sparse regions during training, while MOBA~\cite{lu2025mobamixtureblockattention} employs mixture-of-experts for sparse computation. These methods achieve strong performance but require training or fine-tuning, limiting applicability to pre-trained models.

\subsection{Inference-oriented Sparse Methods}

\textbf{Static Sparse Patterns.} Static methods~\cite{zaheer2020bigbird, Beltagy2020Longformer, xiao2023streamingllm, he2025trianglemixlosslessefficientattention} use predetermined patterns such as local windows, sink tokens, and triangular masks, but cannot adapt to input-specific characteristics.

\textbf{Dynamic Sparse Patterns.} Dynamic methods adaptively compute attention masks during inference. SeerAttention~\cite{gao2024seerattention} employs learned pooling for block selection, MInference~\cite{jiang2024minference} assigns patterns based on offline profiling, FlexPrefill~\cite{lai2025flexprefill} uses divergence metrics for pattern discovery, and XAttention~\cite{xu2025xattention} performs anti-diagonal query sampling with stride aggregation. RRAttention addresses key limitations of these approaches through head round-robin sampling that ensures complete positional coverage and query independence, combined with stride-level aggregation for computational efficiency.

\subsection{Long-context LLM Inference Acceleration}

Beyond sparse attention, orthogonal approaches also accelerate long-context inference. \textbf{Memory-optimized Kernels} like FlashAttention~\cite{dao2022flashattention, dao2023flashattention2} and PagedAttention~\cite{10.1145/3600006.3613165} optimize memory access patterns through kernel fusion and tiling strategies. \textbf{KV Cache Optimization} reduces memory overhead through quantization~\cite{10.5555/3618408.3619696, MLSYS2024_5edb57c0, lin2024qserve}, architectural simplification~\cite{ainslie2023gqatraininggeneralizedmultiquery, deepseekai2024deepseekv3technicalreport}, and cache eviction policies~\cite{zhang2023ho, tang2025razorattention}. These techniques complement sparse attention and can be combined for further acceleration.

\section{Conclusion}
\label{sec:Conclusion}

We present RRAttention, a dynamic sparse attention method achieving optimal balance across five critical dimensions through head round-robin sampling. By rotating query sampling positions across attention heads, RRAttention maintains query independence while enabling efficient global pattern discovery with stride-level aggregation. Extensive experiments demonstrate our method recovers over 99\% of full attention performance while computing only half the attention blocks, achieving 2.4$\times$ speedup at 128K context length. RRAttention consistently outperforms existing dynamic sparse attention methods on both natural language understanding and multimodal video comprehension benchmarks. Future work and extended discussions are in Appendix~\ref{sec:future_work}.

\section{Limitations}

While RRAttention achieves an effective balance across multiple design dimensions through its head round-robin sampling strategy, it is worth noting a boundary condition in extreme configurations. When the stride size $S$ exceeds the number of attention heads, the round-robin sampling scheme cannot guarantee full positional coverage within each stride interval, potentially leaving certain token positions persistently unsampled. This incomplete coverage may affect the completeness of importance estimation for block selection. However, this limitation manifests primarily in impractical configurations: as demonstrated in our stride ablation study (Table~\ref{tab:abl_stride}), excessively large strides already lead to performance degradation due to coarse-grained aggregation that dilutes importance signals. In practical deployments, the recommended moderate stride configurations ($S=8$ or $S=16$) naturally avoid this boundary condition while delivering optimal accuracy-efficiency trade-offs, as these settings ensure sufficient head-stride ratio for comprehensive positional coverage across the sequence.

\bibliography{custom}
\clearpage
\appendix
\section{Detailed Results on HELMET}
\label{tab:app_helmet}

\begin{table*}[!t]
  \centering\footnotesize
  \caption{Detailed performance breakdown across all HELMET task categories using LLaMA-3.1-8B-Instruct.}
  \label{tab:llama_results}
  \setlength{\tabcolsep}{4pt}
  \resizebox{\textwidth}{!}{
  \begin{tabularx}{\textwidth}{
    @{} c |    %
         l |   %
         c |   %
    *{6}{>{\centering\arraybackslash}X} %
         | >{\centering\arraybackslash}X  %
    @{}
  }
    \toprule
    Length & Method & Param (Sparsity) & Recall & RAG & ICL & Cite & Rerank & LongQA & Avg. \\
    \midrule
    \multirow{7}{*}{8k}
      & FullAttention & - (0\%) & 100.0 & 70.12 & 70.96 & 32.89 & 65.82 & 27.21 & 61.17 \\
    \cline{2-10}
      & \multirow{2}{*}{FlexPrefill}
                       & 0.95 (55.29\%) & 95.94 & 65.58 & 68.56 & 23.35 & 54.25 & \textbf{28.44} & 56.02 \\
      &               & 0.99 (26.00\%) & \textbf{100.0} & 68.88 & 70.92 & \textbf{33.23} & 63.88 & 27.47 & 60.73 \\
    \cline{2-10}
      & \multirow{2}{*}{XAttention}
                       & 0.90 (40.30\%) & 97.94 & 68.75 & 70.60 & 29.43 & 65.95 & 26.84 & 59.92 \\
      &               & 0.95 (26.32\%) & 99.25 & 69.25 & 71.00 & 30.38 & \textbf{66.87} & \textbf{28.01} & 60.79 \\
    \cline{2-10}
      & \multirow{2}{*}{\textbf{RRAttention}}
                       & 0.90 (43.48\%) & \textbf{99.75} & \textbf{69.67} & \textbf{71.04} & \textbf{30.33} & \textbf{66.45} & 28.43 & \textbf{60.95} \\
      &               & 0.95 (29.84\%) & \textbf{100.0} & \textbf{69.42} & \textbf{71.12} & \textbf{34.10} & 65.77 & 26.91 & \textbf{61.22} \\
    \midrule
    \multirow{7}{*}{16k}
      & FullAttention & - (0\%) & 100.0 & 67.38 & 75.72 & 24.14 & 51.19 & 33.59 & 58.67 \\
    \cline{2-10}
      & \multirow{2}{*}{FlexPrefill}
                       & 0.95 (61.03\%) & 96.25 & 65.04 & 75.68 & 11.91 & 42.12 & \textbf{35.86} & 54.48 \\
      &               & 0.99 (31.86\%) & \textbf{100.0} & 67.42 & 75.80 & \textbf{21.31} & \textbf{50.89} & 33.50 & 58.15 \\
    \cline{2-10}
      & \multirow{2}{*}{XAttention}
                       & 0.90 (51.68\%) & 97.75 & 67.17 & 76.20 & 19.77 & 48.21 & 33.81 & 57.15 \\
      &               & 0.95 (38.03\%) & 99.25 & 67.42 & 76.16 & 21.17 & 50.74 & 33.89 & 58.10 \\
    \cline{2-10}
      & \multirow{2}{*}{\textbf{RRAttention}}
                       & 0.90 (53.50\%) & \textbf{100.0} & \textbf{67.83} & \textbf{76.36} & \textbf{20.12} & \textbf{51.93} & 34.37 & \textbf{58.43} \\
      &               & 0.95 (39.93\%) & 99.50 & \textbf{67.54} & \textbf{76.28} & 20.50 & 50.84 & \textbf{34.62} & \textbf{58.21} \\
    \midrule
    \multirow{7}{*}{32k}
      & FullAttention & - (0\%) & 99.50 & 65.29 & 78.24 & 11.64 & 42.55 & 43.01 & 56.71 \\
    \cline{2-10}
      & \multirow{2}{*}{FlexPrefill}
                       & 0.95 (66.04\%) & 97.12 & 64.29 & \textbf{79.04} & 8.31 & 29.91 & 39.37 & 53.01 \\
      &               & 0.99 (37.52\%) & \textbf{99.69} & 65.21 & 77.72 & 9.97 & \textbf{42.60} & \textbf{42.15} & \textbf{56.22} \\
    \cline{2-10}
      & \multirow{2}{*}{XAttention}
                       & 0.90 (60.86\%) & 98.12 & 65.25 & 78.68 & \textbf{10.6} & 39.40 & \textbf{41.20} & 55.54 \\
      &               & 0.95 (47.84\%) & 99.00 & 65.29 & \textbf{78.92} & \textbf{10.3} & 39.39 & 40.54 & 55.57 \\
    \cline{2-10}
      & \multirow{2}{*}{\textbf{RRAttention}}
                       & 0.90 (61.51\%) & \textbf{99} & \textbf{65.58} & 78.92 & 10.19 & \textbf{40.86} & 40.60 & \textbf{55.86} \\
      &               & 0.95 (48.49\%) & 99.50 & \textbf{65.50} & 78.72 & 9.95 & 41.04 & \textbf{41.70} & 56.07 \\
    \midrule
    \multirow{7}{*}{64k}
      & FullAttention & - (0\%) & 99.12 & 64.62 & 81.56 & 11.43 & 32.47 & 45.65 & 55.81 \\
    \cline{2-10}
      & \multirow{2}{*}{FlexPrefill}
                       & 0.95 (72.65\%) & 94.50 & 63.88 & 81.84 & 5.82 & 18.21 & 40.40 & 50.77 \\
      &               & 0.99 (47.22\%) & \textbf{99.5} & \textbf{64.92} & 80.68 & \textbf{10.06} & 32.61 & 43.14 & 55.15 \\
    \cline{2-10}
      & \multirow{2}{*}{XAttention}
                       & 0.90 (70.31\%) & 96.19 & 64.33 & \textbf{82.52} & 6.21 & 28.3 & \textbf{44.41} & 53.66 \\
      &               & 0.95 (59.07\%) & 98.56 & 64.50 & \textbf{82.04} & 9.13 & 30.04 & 44.44 & 54.78 \\
    \cline{2-10}
      & \multirow{2}{*}{\textbf{RRAttention}}
                       & 0.90 (70.44\%) & \textbf{98.50} & \textbf{64.50} & 82.36 & \textbf{7.54} & \textbf{28.8} & 43.92 & \textbf{54.27} \\
      &               & 0.95 (59.12\%) & 98.31 & 64.38 & 81.96 & 9.29 & \textbf{33.48} & \textbf{44.59} & \textbf{55.33} \\
    \midrule
    \multirow{7}{*}{128k}
      & FullAttention & - (0\%) & 94.81 & 58.62 & 83.84 & 3.08 & 13.73 & 44.36 & 49.74 \\
    \cline{2-10}
      & \multirow{2}{*}{FlexPrefill}
                       & 0.95 (75.36\%) & \textbf{93.38} & \textbf{59.21} & 85.36 & \textbf{3.07} & 4.40 & \textbf{45.12} & 48.42 \\
      &               & 0.99 (50.54\%) & \textbf{94.94} & 58.67 & 83.32 & 2.83 & \textbf{14.11} & 45.35 & 49.87 \\
    \cline{2-10}
      & \multirow{2}{*}{XAttention}
                       & 0.90 (76.27\%) & 86.12 & 58.67 & \textbf{85.48} & 2.89 & \textbf{10.20} & 43.57 & 47.82 \\
      &               & 0.95 (66.22\%) & 92.50 & 58.96 & \textbf{84.88} & 3.53 & 11.41 & 45.41 & 49.45 \\
    \cline{2-10}
      & \multirow{2}{*}{\textbf{RRAttention}}
                       & 0.90 (76.16\%) & 92.25 & 58.79 & 85.04 & 2.92 & 10.11 & 43.96 & \textbf{48.85} \\
      &               & 0.95 (66.02\%) & 94.75 & \textbf{59.25} & 84.48 & \textbf{3.88} & 13.63 & \textbf{46.24} & \textbf{50.37} \\
    \bottomrule
  \end{tabularx}}
\end{table*}

We evaluate RRAttention on six major task categories from the HELMET benchmark: synthetic recall (Recall), retrieval-augmented generation (RAG), many-shot in-context learning (ICL), generation with citations (Cite), passage re-ranking (Rerank), and long-document question answering (LongQA). Note that we did not evaluate the summarization task as it requires access to GPT API keys for automatic evaluation.

Tables~\ref{tab:llama_results} and~\ref{tab:qwen_results} present comprehensive results across all subtasks at context lengths ranging from 8K to 128K tokens for LLaMA-3.1-8B-Instruct and Qwen2.5-7B-Instruct, respectively. We evaluate two sparsity configurations: a conservative setting ($\gamma=0.99$ and $\tau=0.95$) that prioritizes accuracy retention, and an aggressive setting ($\gamma=0.95$ and $\tau=0.90$) that maximizes computational savings. The following analysis examines performance across different task categories under these two configurations.

\textbf{Global Context Understanding Tasks.} For synthetic recall tasks, RRAttention demonstrates exceptional performance that validates its global evaluation capability. On Llama under the aggressive setting, RRAttention maintains perfect accuracy of 100.0 at 8K and 16K contexts, while FlexPrefill drops to 95.94 and 96.25 respectively. This advantage persists even at extreme 128K context length, where under the conservative setting RRAttention achieves 94.75 compared to FlexPrefill's 94.94 and XAttention's 92.5, while maintaining higher sparsity (66.02\% vs 50.54\% and 66.22\%).

Long-document question answering represents another critical test requiring comprehensive context understanding. On Qwen at 128K context under the aggressive setting, RRAttention achieves 42.15, substantially outperforming XAttention's 40.92 and dramatically surpassing FlexPrefill's 20.68, with comparable sparsity (71.29\% vs 71.25\% and 71.58\%). Under the conservative setting at the same context length, RRAttention achieves 42.83 versus XAttention's 42.46 and FlexPrefill's 35.46 (60.97\% vs 60.92\% and 48.20\% sparsity). At 64K context under the conservative setting, RRAttention achieves 47.20 versus XAttention's 46.85 and FlexPrefill's 43.99 (52.99\% vs 53.36\% and 40.89\% sparsity). This consistent advantage validates that RRAttention's head round-robin sampling strategy effectively captures diverse attention patterns across extended contexts.

\textbf{Information Retrieval Tasks.} For retrieval-augmented generation tasks, RRAttention consistently outperforms baselines by better capturing retrieval-relevant patterns. On Qwen at 32K context under the aggressive setting, RRAttention achieves 55.92, compared to XAttention's 55.04 and FlexPrefill's 48.33 (59.66\% vs 59.25\% and 62.66\% sparsity). At 128K context under the aggressive setting, RRAttention achieves 47.46 versus XAttention's 46.88 and FlexPrefill's 34.04 (71.29\% vs 71.25\% and 71.58\% sparsity), demonstrating robust performance even under extreme context lengths.

In many-shot in-context learning tasks, all methods maintain relatively high performance with RRAttention showing competitive results. On Llama under the aggressive setting, RRAttention achieves the highest ICL scores at shorter contexts: 71.04 at 8K and 76.36 at 16K (43.48\% and 53.50\% sparsity respectively). The gap narrows at longer contexts where all methods approach similar performance levels (around $84{\sim}85$ at 128K), indicating that ICL tasks have more concentrated attention patterns that are easier for sparse methods to capture.

\textbf{Fine-grained Attribution Tasks.} Citation generation and passage re-ranking tasks present unique challenges for sparse attention methods. Citation generation proves most challenging for all methods, with substantial performance degradation at longer contexts. On Llama at 64K context under the conservative setting, RRAttention achieves 9.29, compared to XAttention's 9.13 and FlexPrefill's 10.06 (59.12\% vs 59.07\% and 47.22\% sparsity). At 128K under the conservative setting, even the best sparse methods only partially recover FullAttention performance (RRAttention: 3.88 vs FullAttention: 3.08, with 66.02\% sparsity), suggesting that precise attribution requiring fine-grained attention to specific positions remains inherently challenging.

Passage re-ranking exhibits context-dependent behavior. At 64K context on Llama under the conservative setting, RRAttention achieves 33.48, substantially outperforming XAttention's 30.04 and nearly matching FlexPrefill's 32.61 despite achieving much higher sparsity (59.12\% vs 59.07\% and 47.22\%). However, performance degrades significantly at 128K where all sparse methods struggle, highlighting the difficulty of maintaining ranking signals under extreme sparsification.

\textbf{Sparsity-Accuracy Trade-off Analysis.} RRAttention achieves superior sparsity-accuracy trade-offs across settings. Under the aggressive setting on Qwen at 64K context, RRAttention achieves 44.62, compared to XAttention's 43.35 and FlexPrefill's 36.39 (64.78\% vs 65.07\% and 66.09\% sparsity), demonstrating significantly better pattern selection at similar sparsity levels. Under the conservative setting, RRAttention achieves 45.17, outperforming XAttention's 44.49 and substantially exceeding FlexPrefill's 43.01 (52.99\% vs 53.56\% and 40.89\% sparsity).

The consistent performance gains across both Llama and Qwen models demonstrate RRAttention's generalizability. While absolute scores differ between models due to architectural differences, the relative advantages of RRAttention remain consistent, validating that our head round-robin sampling strategy captures fundamental attention properties rather than exploiting model-specific characteristics.

\begin{table*}[!t]
  \centering\footnotesize
  \caption{Detailed performance breakdown across all HELMET task categories using Qwen2.5-7B-Instruct.}
  \label{tab:qwen_results}
  \setlength{\tabcolsep}{4pt}
  \resizebox{\textwidth}{!}{
  \begin{tabularx}{\textwidth}{
    @{} c |    %
         l |   %
         c |   %
    *{6}{>{\centering\arraybackslash}X} %
         | >{\centering\arraybackslash}X  %
    @{}
  }
    \toprule
    Length & Method & Param (Sparsity) & Recall & RAG & ICL & Cite & Rerank & LongQA & Avg. \\
    \midrule
    \multirow{7}{*}{8k}
      & FullAttention & - (0\%) & 78.88 & 60.75 & 69.72 & 27.29 & 57.84 & 34.04 & 54.75 \\
    \cline{2-10}
      & \multirow{2}{*}{FlexPrefill}
                       & 0.95 (56.61\%) & 59.56 & 52.08 & 67.84 & 13.68 & 39.78 & 31.70 & 44.11 \\
      &               & 0.99 (31.01\%) & 76.62 & 58.29 & \textbf{70.12} & 25.21 & 52.86 & \textbf{35.99} & 53.18 \\
    \cline{2-10}
      & \multirow{2}{*}{XAttention}
                       & 0.90 (46.35\%) & 72.69 & 59.38 & \textbf{69.44} & \textbf{25.59} & 57.16 & 30.96 & 52.53 \\
      &               & 0.95 (33.90\%) & 77.12 & \textbf{59.88} & 69.64 & \textbf{26.76} & \textbf{57.89} & 32.13 & 53.90 \\
    \cline{2-10}
    & \multirow{2}{*}{\textbf{RRAttention}}
                       & 0.90 (48.14\%) & \textbf{77.62} & \textbf{59.50} & 69.08 & 22.88 & \textbf{58.39} & \textbf{32.74} & \textbf{53.37} \\
      &               & 0.95 (35.79\%) & \textbf{79.31} & 59.83 & 69.40 & 25.45 & 58.21 & 33.09 & \textbf{54.22} \\
    \midrule
    \multirow{7}{*}{16k}
      & FullAttention & - (0\%) & 77.81 & 59.46 & 74.32 & 16.94 & 37.96 & 37.44 & 50.66 \\
    \cline{2-10}
      & \multirow{2}{*}{FlexPrefill}
                       & 0.95 (59.12\%) & 54.25 & 50.21 & 72.80 & 8.10 & 23.59 & 34.56 & 40.59 \\
      &               & 0.99 (32.80\%) & 74.12 & 57.21 & 73.64 & 13.61 & 37.55 & 38.17 & 49.05 \\
    \cline{2-10}
      & \multirow{2}{*}{XAttention}
                       & 0.90 (53.40\%) & 65.25 & \textbf{59.67} & \textbf{74.72} & 12.06 & \textbf{37.06} & 37.54 & 47.72 \\
      &               & 0.95 (40.88\%) & 72.62 & \textbf{60.08} & \textbf{74.48} & \textbf{16.39} & \textbf{38.07} & 36.95 & 49.77 \\
    \cline{2-10}
      & \multirow{2}{*}{\textbf{RRAttention}}
                       & 0.90 (54.25\%) & \textbf{74.00} & 59.21 & 74.12 & \textbf{13.96} & 35.97 & \textbf{38.58} & \textbf{49.31} \\
      &               & 0.95 (41.61\%) & \textbf{76.31} & 59.17 & 74.24 & 15.38 & 36.62 & \textbf{38.39} & \textbf{50.02} \\
    \midrule
    \multirow{7}{*}{32k}
      & FullAttention & - (0\%) & 76.25 & 56.17 & 76.32 & 13.87 & 25.39 & 42.42 & 48.40 \\
    \cline{2-10}
      & \multirow{2}{*}{FlexPrefill}
                       & 0.95 (62.66\%) & 57.75 & 48.33 & 76.32 & 7.27 & 15.13 & 38.82 & 40.60 \\
      &               & 0.99 (36.48\%) & \textbf{75.81} & 55.58 & 74.92 & 11.86 & 21.86 & 38.82 & 46.48 \\
    \cline{2-10}
      & \multirow{2}{*}{XAttention}
                       & 0.90 (59.25\%) & 65.75 & 55.04 & \textbf{76.44} & \textbf{12.77} & 20.52 & \textbf{43.82} & 45.72 \\
      &               & 0.95 (47.04\%) & 67.62 & 55.79 & 76.24 & \textbf{13.99} & 25.33 & \textbf{42.77} & 46.96 \\
    \cline{2-10}
      & \multirow{2}{*}{\textbf{RRAttention}}
                       & 0.90 (59.66\%) & \textbf{73.69} & \textbf{55.92} & 76.24 & 12.12 & \textbf{24.20} & 41.91 & \textbf{47.35} \\
      &               & 0.95 (47.45\%) & 74.00 & \textbf{55.96} & \textbf{76.52} & 12.66 & \textbf{26.51} & 41.89 & \textbf{47.92} \\
    \midrule
    \multirow{7}{*}{64k}
      & FullAttention & - (0\%) & 67.25 & 55.21 & 77.00 & 11.58 & 13.72 & 45.16 & 44.98 \\
    \cline{2-10}
      & \multirow{2}{*}{FlexPrefill}
                       & 0.95 (66.09\%) & 51.94 & 44.04 & 76.60 & 4.20 & 7.41 & 34.13 & 36.39 \\
      &               & 0.99 (40.89\%) & 62.62 & 53.79 & 75.44 & 6.92 & 15.30 & 43.99 & 43.01 \\
    \cline{2-10}
      & \multirow{2}{*}{XAttention}
                       & 0.90 (65.07\%) & 55.12 & 55.21 & 77.12 & 8.96 & \textbf{16.19} & \textbf{47.49} & 43.35 \\
      &               & 0.95 (53.56\%) & 60.94 & \textbf{55.33} & \textbf{77.04} & 10.85 & \textbf{15.92} & 46.85 & 44.49 \\
    \cline{2-10}
      & \multirow{2}{*}{\textbf{RRAttention}}
                       & 0.90 (64.78\%) & \textbf{63.38} & \textbf{55.29} & \textbf{77.56} & \textbf{9.51} & 15.28 & 46.72 & \textbf{44.62} \\
      &               & 0.95 (52.99\%) & \textbf{65.12} & 54.96 & 76.80 & \textbf{11.22} & 15.73 & \textbf{47.20} & \textbf{45.17} \\
    \midrule
    \multirow{7}{*}{128k}
      & FullAttention & - (0\%) & 50.25 & 47.08 & 79.24 & 6.54 & 8.55 & 45.87 & 39.59 \\
    \cline{2-10}
      & \multirow{2}{*}{FlexPrefill}
                       & 0.95 (71.58\%) & 38.62 & 34.04 & 77.92 & 2.13 & 4.54 & 20.68 & 29.66 \\
      &               & 0.99 (48.20\%) & 45.94 & 43.79 & 78.12 & 3.70 & 6.13 & 35.46 & 35.52 \\
    \cline{2-10}
      & \multirow{2}{*}{XAttention}
                       & 0.90 (71.25\%) & 43.12 & 46.88 & \textbf{79.96} & \textbf{5.58} & 5.78 & 40.92 & 37.04 \\
      &               & 0.95 (60.92\%) & 45.25 & \textbf{48.08} & 79.40 & \textbf{5.88} & 7.48 & 42.46 & 38.09 \\
    \cline{2-10}
      & \multirow{2}{*}{\textbf{RRAttention}}
                       & 0.90 (71.29\%) & \textbf{48.31} & \textbf{47.46} & 79.68 & 4.72 & \textbf{8.56} & \textbf{42.15} & \textbf{38.48} \\
      &               & 0.95 (60.97\%) & \textbf{47.44} & 47.33 & \textbf{79.80} & 5.82 & \textbf{7.85} & \textbf{42.83} & \textbf{38.51} \\
    \bottomrule
  \end{tabularx}}
\end{table*}

\section{Additional Model Experiments}
\label{sec:appendix_additional_models}

To further validate the generalizability of RRAttention across diverse model families and scales, we conduct additional experiments on two models: (i) \textbf{Yi-9B-200K}~\cite{ai2025yiopenfoundationmodels}, which supports up to 200K tokens context length through lightweight continual pretraining; (ii) \textbf{Qwen3-30B-A3B-Instruct-2507}~\cite{qwen3technicalreport}, a larger-scale Mixture-of-Experts model supporting up to 256K context length, which allows us to evaluate whether RRAttention's advantages transfer to models with significantly more parameters. We evaluate all methods on the HELMET benchmark across context lengths from 8K to 128K tokens using the same experimental settings as our main experiments.

\begin{table*}[!t]
  \centering
  \caption{Performance comparison on additional model families across the HELMET benchmark.}
  \vskip -0.1in
  \label{tab:additional_models}
  \resizebox{0.9\textwidth}{!}{
  \begin{tabular}{c | l | c | c c c c c | c}
    \toprule
    \textbf{Model} & \textbf{Methods} & \textbf{Avg. Sparsity(↑)} & \textbf{8K} & \textbf{16K} & \textbf{32K} & \textbf{64K} & \textbf{128K} & \textbf{Avg.(↑)} \\
    \midrule
    \multirow{7}{*}{Yi-9B-200K}
      & FullAttention & - (0\%) & 50.33 & 48.87 & 46.11 & 43.83 & 41.07 & 46.04 \\
    \cline{2-9}
      & \multirow{2}{*}{FlexPrefill}
                        & 0.95 (65.98\%) & 44.99 & 44.51 & 43.27 & 41.31 & 38.73 & 42.56 \\
      &                 & 0.99 (40.16\%) & 50.87 & \textbf{50.01} & 46.00 & 43.87 & 40.92 & 46.33 \\
    \cline{2-9}
      & \multirow{2}{*}{XAttention}
                        & 0.90 (63.85\%) & 50.28 & 47.79 & 44.17 & 43.31 & 40.19 & 45.15 \\
      &                 & 0.95 (50.80\%) & 50.41 & 48.51 & 44.73 & 43.69 & \textbf{41.17} & 45.70 \\
    \cline{2-9}
      & \multirow{2}{*}{\textbf{RRAttention}}
                        & 0.90 (64.05\%) & \textbf{50.79} & \textbf{48.58} & \textbf{46.48} & \textbf{43.89} & \textbf{40.77} & \textbf{46.10} \\
      &                 & 0.95 (51.86\%) & \textbf{50.95} & 49.45 & \textbf{47.13} & \textbf{44.12} & 41.06 & \textbf{46.54} \\
    \midrule
    \multirow{7}{*}{\shortstack{Qwen3-30B-\\A3B-Instruct}}
      & FullAttention & - (0\%) & 67.09 & 67.42 & 66.78 & 65.73 & 62.80 & 65.96 \\
    \cline{2-9}
      & \multirow{2}{*}{FlexPrefill}
                        & 0.95 (58.42\%) & 64.08 & 64.11 & 63.15 & 61.64 & 59.30 & 62.46 \\
      &                 & 0.97 (48.22\%) & 66.15 & 65.61 & 64.97 & 63.20 & 60.80 & 64.15 \\
    \cline{2-9}
      & \multirow{2}{*}{XAttention}
                        & 0.90 (55.20\%) & 66.78 & \textbf{67.19} & 65.25 & \textbf{63.12} & 60.04 & 64.48 \\
      &                 & 0.95 (41.96\%) & 66.51 & 67.22 & 66.38 & 63.75 & 61.17 & 65.01 \\
    \cline{2-9}
      & \multirow{2}{*}{\textbf{RRAttention}}
                        & 0.90 (56.26\%) & \textbf{67.01} & 66.60 & \textbf{65.71} & 62.97 & \textbf{60.70} & \textbf{64.60} \\
      &                 & 0.95 (43.01\%) & \textbf{66.96} & \textbf{67.63} & \textbf{66.55} & \textbf{64.46} & \textbf{61.66} & \textbf{65.45} \\
    \bottomrule
  \end{tabular}}
\end{table*}

Table~\ref{tab:additional_models} presents the results on both additional model families. On Yi-9B-200K, RRAttention generally achieves the highest accuracy among all sparse attention methods. At the conservative setting ($\tau=0.95$ and $\gamma=0.99$), RRAttention achieves 46.54, compared to XAttention's 45.70 and FlexPrefill's 46.33 (51.86\% vs 50.80\% and 40.16\% sparsity). Notably, RRAttention even surpasses FullAttention accuracy (46.54 vs 46.04) while computing approximately half of the attention blocks, suggesting that selective attention can sometimes act as beneficial regularization by filtering out noisy or irrelevant context. The accuracy-sparsity trade-off advantage of RRAttention is even more pronounced at the aggressive setting ($\tau=0.90$ and $\gamma=0.95$). RRAttention maintains 46.10, substantially outperforming XAttention's 45.15 and FlexPrefill's 42.56 (64.05\% vs 63.85\% and 65.98\% sparsity).

On the larger-scale Qwen3-30B-A3B-Instruct model, RRAttention's advantages transfer effectively. At the conservative setting ($\tau=0.95$ and $\gamma=0.97$), RRAttention achieves 65.45, compared to XAttention's 65.01 and FlexPrefill's 64.15 (43.01\% vs 41.96\% and 48.22\% sparsity). RRAttention recovers 99.2\% of FullAttention performance (65.45 vs 65.96) while computing less than half of the attention blocks. The pattern holds at the aggressive setting ($\tau=0.90$ and $\gamma=0.95$), where RRAttention achieves 64.60, compared to XAttention's 64.48 and FlexPrefill's 62.46 (56.26\% vs 55.20\% and 58.42\% sparsity).

These results across four diverse model families (LLaMA-3.1-8B, Qwen2.5-7B, Yi-9B-200K, and Qwen3-30B-A3B) demonstrate that RRAttention more effectively identifies critical attention patterns—achieving higher sparsity than XAttention while maintaining better accuracy, and preserving significantly better accuracy than FlexPrefill at comparable sparsity levels. The consistent improvements confirm that our head round-robin sampling strategy generalizes well across different architectures, model scales, and context length capabilities.

\section{Inference Time Breakdown}
\label{sec:appendix_time_breakdown}

To provide transparency into the computational cost of each component in RRAttention, we conduct a detailed timing analysis across different context lengths. All experiments are performed on NVIDIA H100 GPU using LLaMA-3.1-8B-Instruct with the conservative setting ($\tau=0.95$). We measure the execution time of four stages: (1) query sampling with head round-robin strategy, (2) stride-level importance estimation, (3) block-level selection via top-$\tau$ thresholding, and (4) sparse attention computation.

\begin{table*}[h]
\centering
\caption{RRAttention inference time breakdown by stage (LLaMA-3.1-8B-Instruct, H100 GPU, $\tau=0.95$).}
\label{tab:time_breakdown}
\resizebox{\textwidth}{!}{
\begin{tabular}{lccccc}
\toprule
Context & Query Sampling (ms) & Stride-level Estimation (ms) & Block Selection (ms) & Sparse Attention (ms) & Pattern Search \% \\
\midrule
8K & 0.008 (0.01\%) & 26.67 (21.2\%) & 7.72 (6.1\%) & 90.62 (72.0\%) & 27.3\% \\
32K & 0.012 (0.02\%) & 104.32 (13.8\%) & 34.59 (4.6\%) & 617.86 (81.7\%) & 18.3\% \\
64K & 0.019 (0.01\%) & 296.98 (13.8\%) & 94.34 (4.4\%) & 1720.33 (79.8\%) & 18.5\% \\
96K & 0.027 (0.01\%) & 785.61 (15.4\%) & 214.29 (4.2\%) & 4154.46 (81.4\%) & 19.4\% \\
128K & 0.036 (0.01\%) & 956.21 (15.4\%) & 312.85 (5.0\%) & 5221.27 (84.2\%) & 19.5\% \\
\bottomrule
\end{tabular}
}
\end{table*}

As shown in Table~\ref{tab:time_breakdown}, the query sampling stage incurs negligible overhead ($<$0.04ms, $<$0.02\% of total time) across all context lengths, as it involves only simple indexing operations with $O(H \cdot L/S)$ complexity. The stride-level importance estimation dominates the pattern search phase (77.5\%-96.7\% of pattern discovery time), scaling from 26.67ms at 8K to 956.21ms at 128K, which reflects the $O(L^2/S^2)$ complexity of our approach. The block-level selection stage remains lightweight (4.2\%-6.1\% of pattern search time) due to efficient GPU-accelerated sorting. The sparse attention computation constitutes the dominant overall cost (72.0\%-84.2\% of total time), which is unavoidable and shared across all sparse attention methods. Notably, the combined pattern search overhead (stages 1-3) accounts for only 18-19\% of total inference time at long contexts ($\geq$32K), demonstrating that RRAttention's $O(L^2/S^2)$ complexity enables practical dynamic pattern discovery with minimal overhead compared to the actual sparse attention computation.

\section{Block Selection Accuracy Analysis}
\label{appendix:block_selection}

\textbf{Problem Formulation.} Let $\mathbf{A} \in \mathbb{R}^{L \times L}$ denote the full attention matrix where $\mathbf{A}_{i,j}$ represents the softmax attention score from query position $i$ to key position $j$. For each query position $i$, we define the ground truth important key set as:
\begin{equation}
\label{true_attention}
\mathcal{K}_i^* = \left\{j \mid j \in \underset{\mathcal{S} \subseteq [L]}{\arg\min} \left\{|\mathcal{S}| : \sum_{k \in \mathcal{S}} \mathbf{A}_{i,k} \geq \tau\right\}\right\}
\end{equation}
where $\tau = 0.95$ is the cumulative attention threshold. This represents the minimal set of key positions contributing 95\% of the attention mass for query $i$.

For block-based sparse attention methods operating at block size $B$, let $\mathcal{B}_i \subseteq \{0, 1, \ldots, \lfloor L/B \rfloor - 1\}$ denote the set of selected key blocks for query block containing position $i$. The predicted important key set is:
\begin{equation}
\mathcal{K}_i = \bigcup_{b \in \mathcal{B}_i} \{j \mid bB \leq j < (b+1)B\}
\end{equation}

We evaluate selection accuracy using:
\begin{align}
\text{Precision} &= \frac{1}{L}\sum_{i=1}^{L} \frac{|\mathcal{K}_i \cap \mathcal{K}_i^*|}{|\mathcal{K}_i|} \\
\text{Recall} &= \frac{1}{L}\sum_{i=1}^{L} \frac{|\mathcal{K}_i \cap \mathcal{K}_i^*|}{|\mathcal{K}_i^*|} \\
\text{F1} &= 2 \cdot \frac{\text{Precision} \cdot \text{Recall}}{\text{Precision} + \text{Recall}}
\end{align}

\textbf{Experimental Setup.} We conduct this analysis on the HELMET benchmark using LLaMA-3.1-8B-Instruct across multiple context lengths. We compare XAttention-0.95 and RRAttention-0.95, both configured with 95\% sparsity threshold at block granularity. Block size is set to $B=128$ following our main experiments. Results are averaged across all evaluated context lengths.

\begin{table*}[!t]
  \centering
  \caption{Block selection accuracy averaged across context lengths on HELMET subtasks. All values are percentages.}
  \label{tab:abl_selection_accuracy}
  \resizebox{\textwidth}{!}{
  \begin{tabular}{l | c c c | c c c | c c c | c c c | c c c | c c c}
    \toprule
    \multirow{2}{*}{\textbf{Method}} & \multicolumn{3}{c|}{\textbf{Recall}} & \multicolumn{3}{c|}{\textbf{Cite}} & \multicolumn{3}{c|}{\textbf{Rerank}} & \multicolumn{3}{c|}{\textbf{Longqa}} & \multicolumn{3}{c|}{\textbf{ICL}} & \multicolumn{3}{c}{\textbf{RAG}} \\
    \cmidrule(lr){2-4} \cmidrule(lr){5-7} \cmidrule(lr){8-10} \cmidrule(lr){11-13} \cmidrule(lr){14-16} \cmidrule(lr){17-19}
    & Prec. & Rec. & F1 & Prec. & Rec. & F1 & Prec. & Rec. & F1 & Prec. & Rec. & F1 & Prec. & Rec. & F1 & Prec. & Rec. & F1 \\
    \midrule
    XAttention & 12.48 & \textbf{93.35} & 26.81 & 13.63 & \textbf{91.79} & 22.37 & 10.30 & \textbf{93.35} & 17.91 & 13.63 & \textbf{91.79} & 22.37 & 15.11 & \textbf{93.27} & 57.07 & 10.95 & \textbf{91.61} & 52.44 \\
    RRAttention & \textbf{13.05} & 93.05 & \textbf{27.58} & \textbf{14.16} & 91.64 & \textbf{22.92} & \textbf{10.78} & 93.05 & \textbf{18.65} & \textbf{14.16} & 91.64 & \textbf{22.92} & \textbf{15.63} & 92.90 & \textbf{57.32} & \textbf{11.44} & 91.27 & \textbf{52.69} \\
    \bottomrule
  \end{tabular}}
\end{table*}

\textbf{Results and Analysis.} Table~\ref{tab:abl_selection_accuracy} presents the block selection accuracy averaged across context lengths for all HELMET subtasks. Several key observations emerge from these results.

First, both methods achieve consistently high recall rates (91-93\% across all tasks), with RRAttention's recall within 0.3\% of XAttention. This demonstrates that stride-level aggregation effectively captures the vast majority of important attention patterns regardless of the query sampling strategy, confirming that neither approach sacrifices pattern coverage capability.

Second, RRAttention demonstrates systematically higher precision than XAttention across all six tasks. The precision improvements range from 0.48\% (Rerank) to 0.57\% (Recall), averaging 0.52\% across tasks. This represents approximately 4-5\% relative reduction in false positive block selections. The consistency of this improvement across diverse task types—from simple retrieval (Recall, RAG) to complex reasoning (Longqa, Cite) to many-shot learning (ICL)—validates that head round-robin sampling's superiority in pattern discovery generalizes across different attention pattern characteristics.

Third, the F1 scores reflect RRAttention's superior overall selection accuracy, with improvements ranging from 0.25 (RAG) to 0.77 (Recall) F1 points across tasks. Tasks requiring precise attention pattern identification show the most substantial gains: Recall (+0.77), Rerank (+0.74), and Cite/Longqa (+0.55 each). Even in the ICL task with already high F1 scores (57.07), RRAttention achieves further improvement (+0.25), demonstrating robustness across varying task difficulties.

\textbf{Conclusion.} These results provide quantitative evidence that RRAttention's head round-robin sampling strategy enables more accurate identification of important attention patterns compared to fixed-position sampling approaches. The consistently higher precision (averaging +0.52\%) with maintained recall demonstrates that head round-robin sampling reduces false positive selections while preserving pattern coverage. The improved pattern discovery accuracy—manifested through higher F1 scores across all six diverse tasks—validates our core design principle: ensuring complete positional coverage across attention heads within each stride leads to fundamentally better importance estimation. This enhanced block selection accuracy directly translates to the superior end-to-end task performance observed in our main experiments, confirming that accurate pattern discovery is crucial for effective dynamic sparse attention.

\section{Attention Pattern Visualization}
\label{sec:appendix_visualization}

To provide intuitive understanding of how RRAttention identifies critical attention patterns, we visualize the selected attention blocks at different context lengths. For each visualization, we compare the ground truth important blocks ($\tau=0.95$, computed from FullAttention scores as defined in Equation~\ref{true_attention}) with the blocks selected by RRAttention ($\tau=0.95$). We present representative attention heatmaps from different layers and heads across context lengths of 16K, 32K, 64K, and 128K tokens.

\begin{figure*}[htp]
    \centering
    \subfloat[FullAttention]{
        \includegraphics[width=0.24\textwidth]{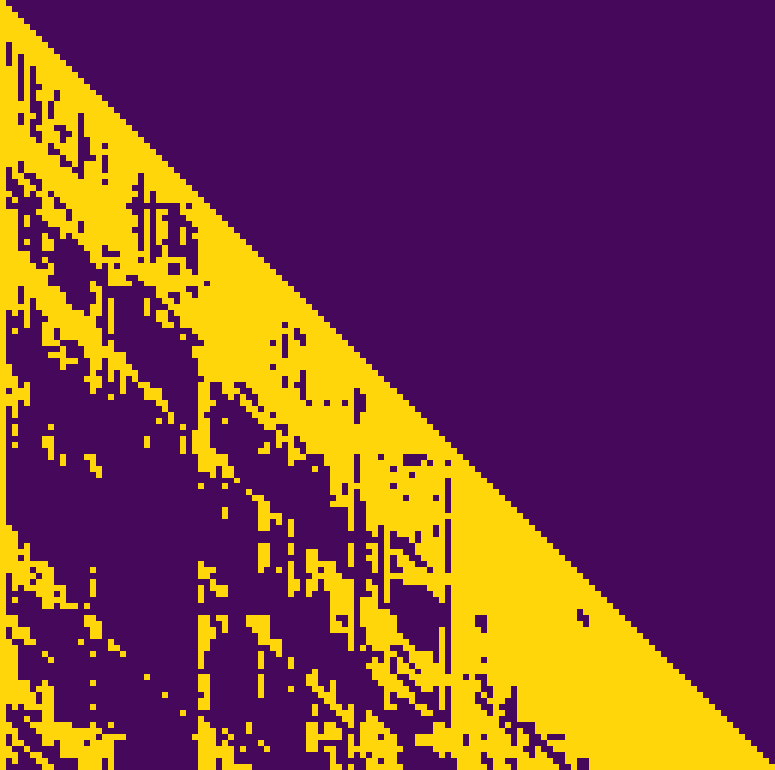}
    }%
    \subfloat[RRAttention]{
        \includegraphics[width=0.24\textwidth]{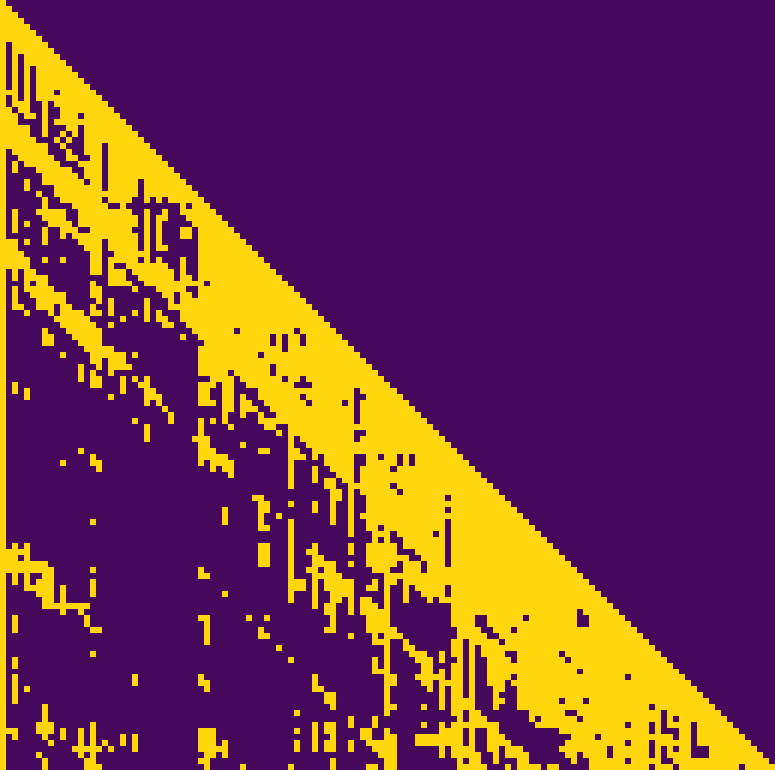}
    }%
    \subfloat[FullAttention]{
        \includegraphics[width=0.24\textwidth]{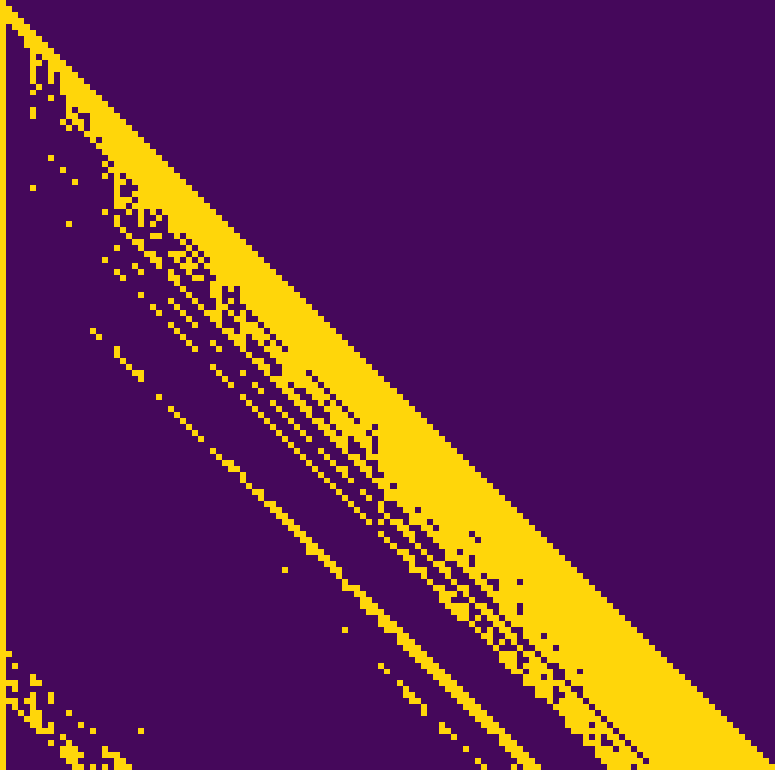}
    }%
    \subfloat[RRAttention]{
        \includegraphics[width=0.24\textwidth]{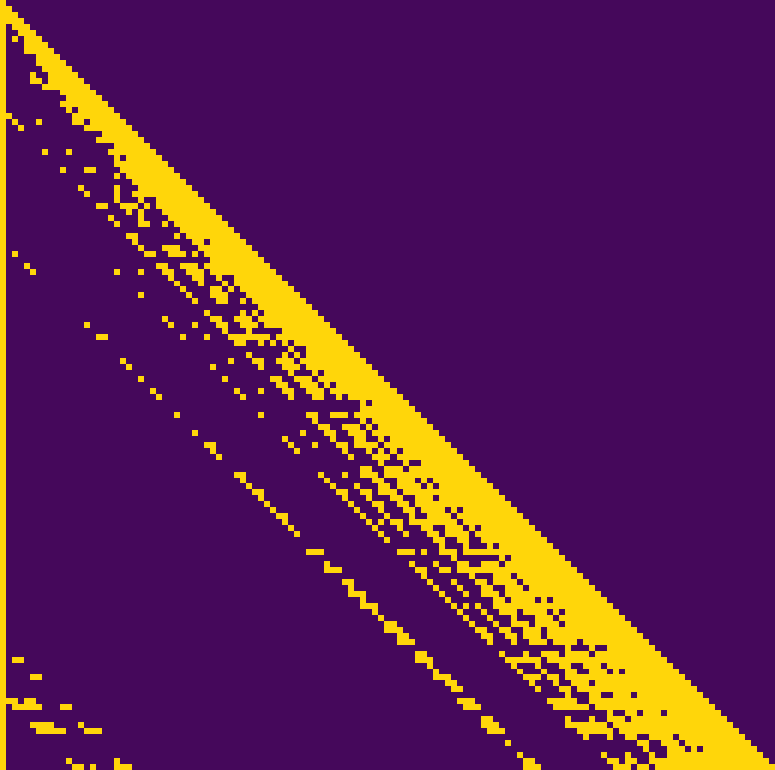}
    }
    \caption{Attention pattern visualization at 16K context length. Each pair shows FullAttention ground truth (left) and RRAttention selection (right).}
    \label{fig:vis_16k}
\end{figure*}
\begin{figure*}[htp]
    \centering
    \subfloat[FullAttention]{
        \includegraphics[width=0.24\textwidth]{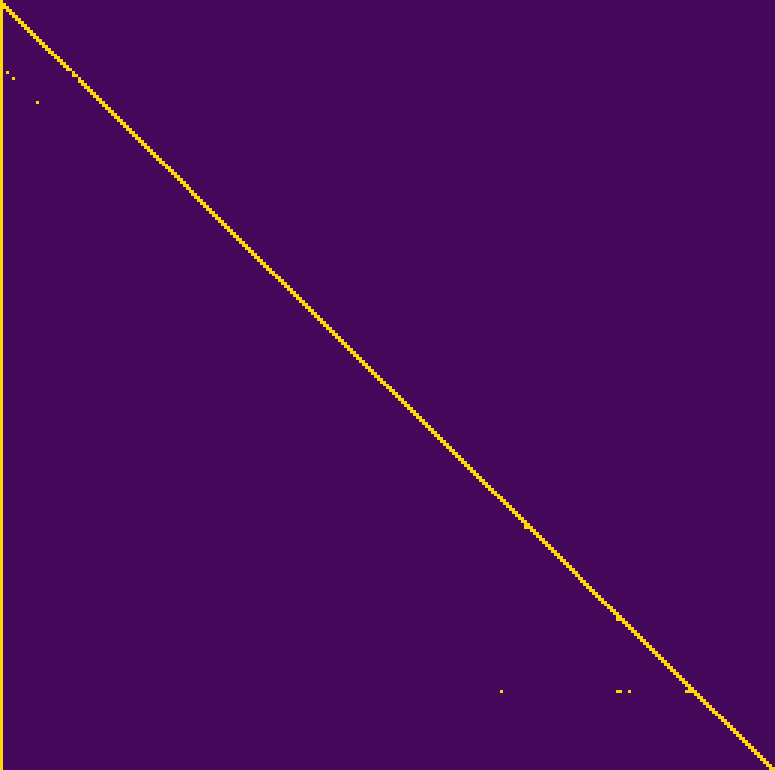}
    }%
    \subfloat[RRAttention]{
        \includegraphics[width=0.24\textwidth]{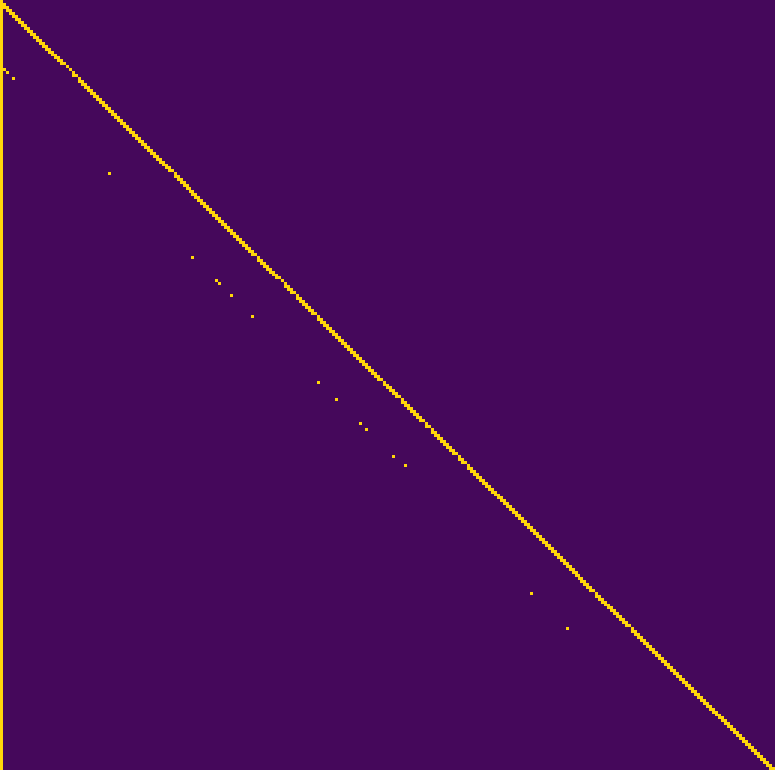}
    }%
    \subfloat[FullAttention]{
        \includegraphics[width=0.24\textwidth]{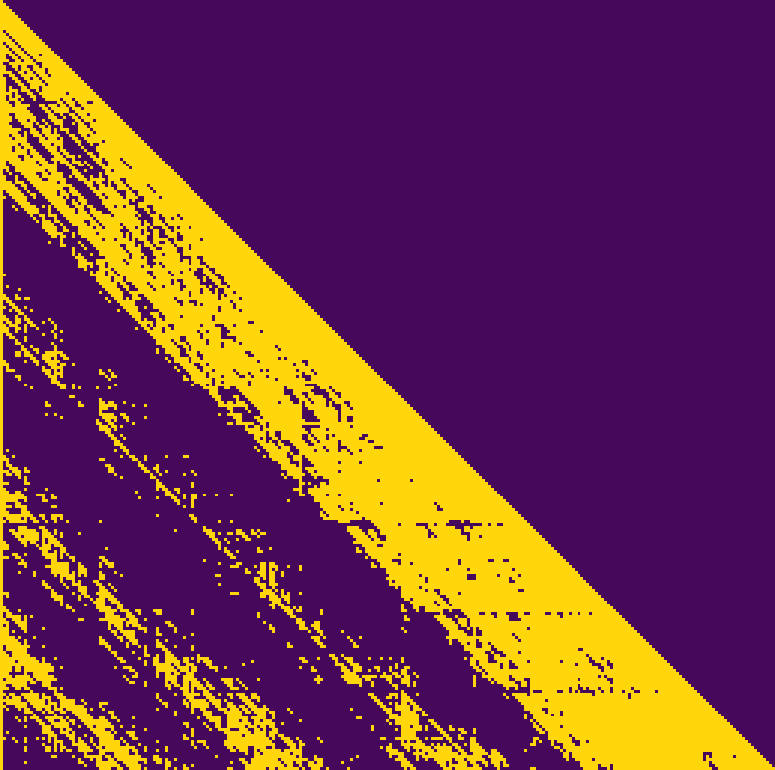}
    }%
    \subfloat[RRAttention]{
        \includegraphics[width=0.24\textwidth]{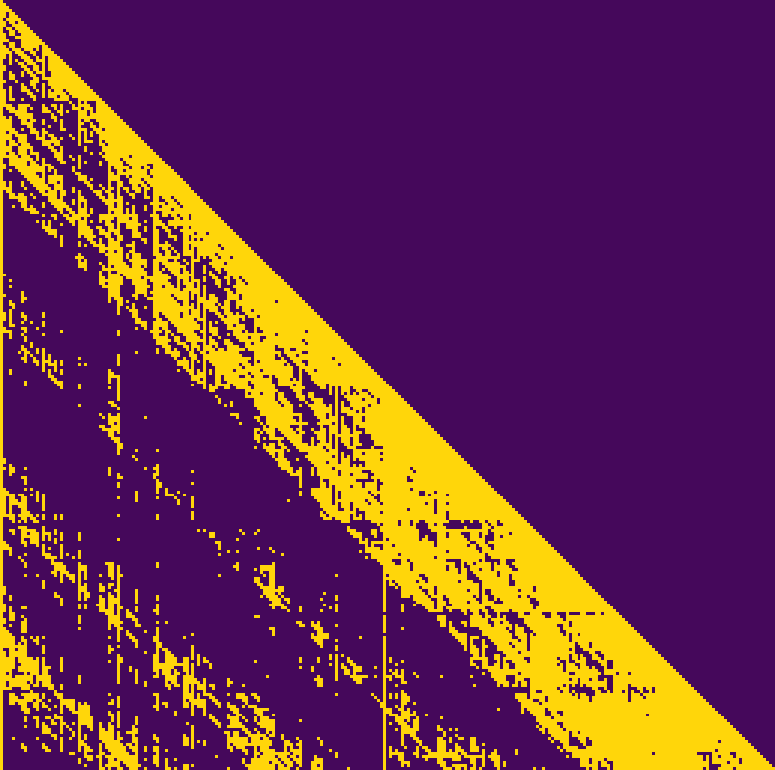}
    }
    \caption{Attention pattern visualization at 32K context length. Each pair shows FullAttention ground truth (left) and RRAttention selection (right).}
    \label{fig:vis_32k}
\end{figure*}
\begin{figure*}[htp]
    \centering
    \subfloat[FullAttention]{
        \includegraphics[width=0.24\textwidth]{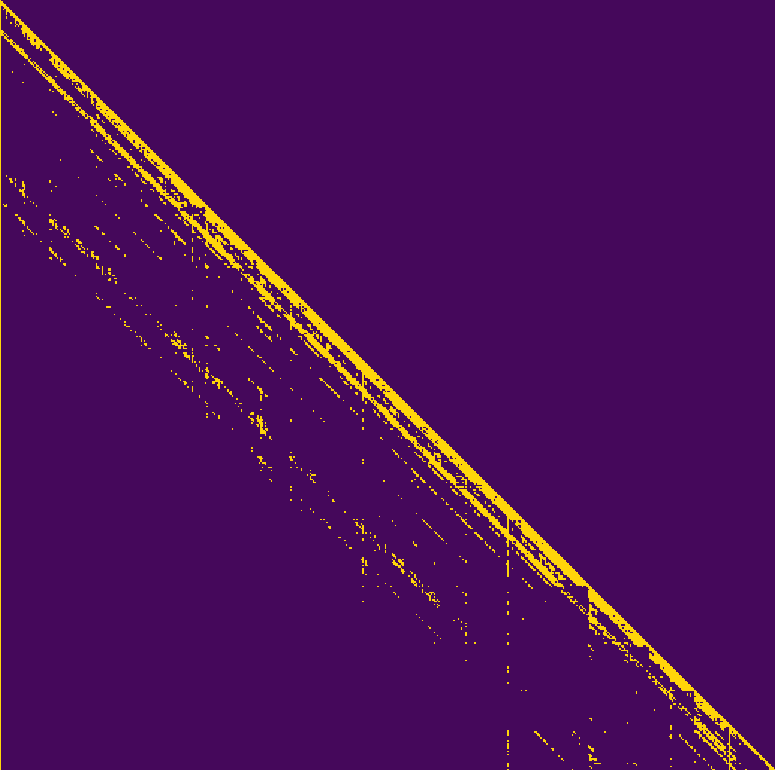}
    }%
    \subfloat[RRAttention]{
        \includegraphics[width=0.24\textwidth]{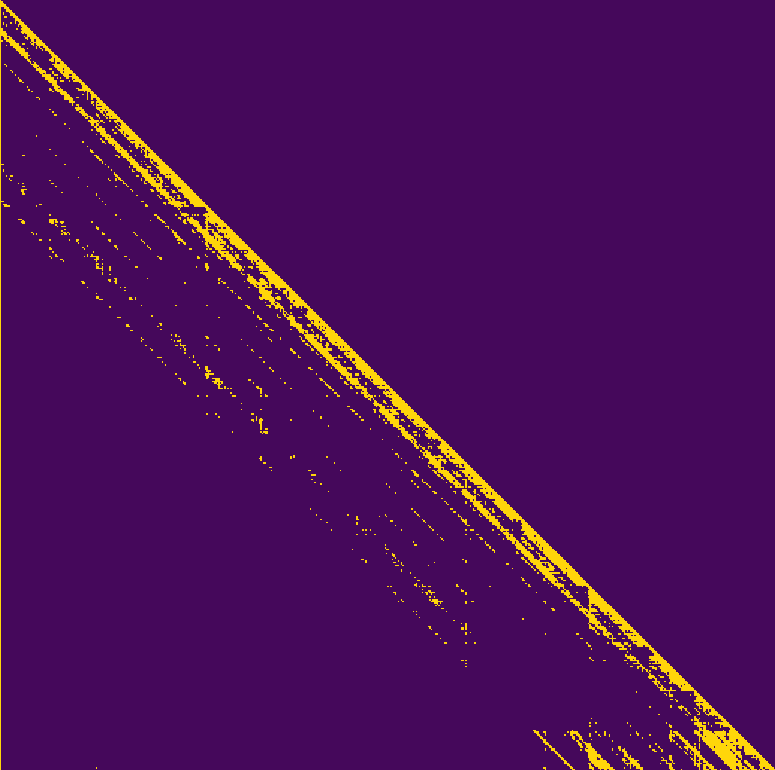}
    }%
    \subfloat[FullAttention]{
        \includegraphics[width=0.24\textwidth]{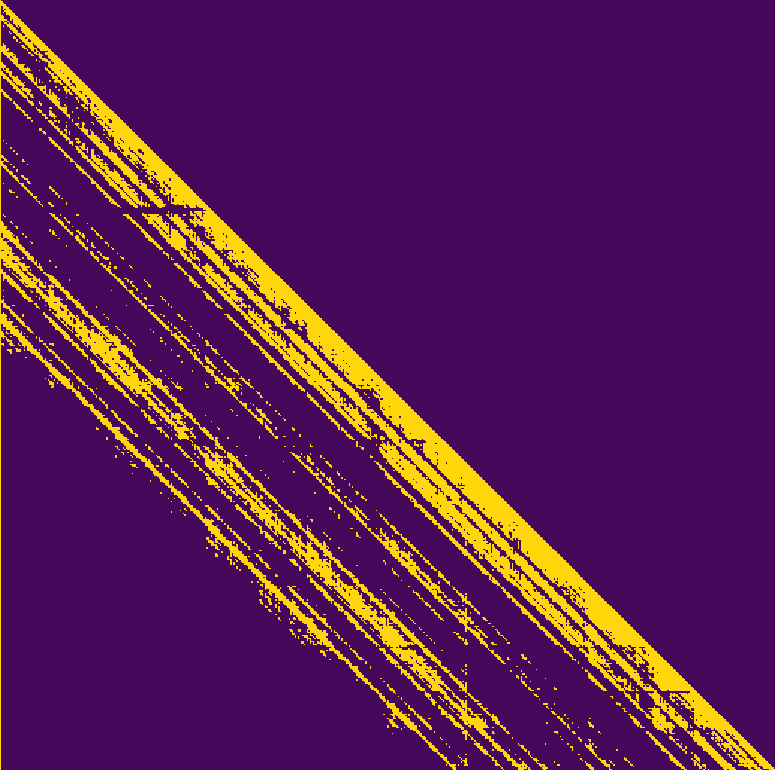}
    }%
    \subfloat[RRAttention]{
        \includegraphics[width=0.24\textwidth]{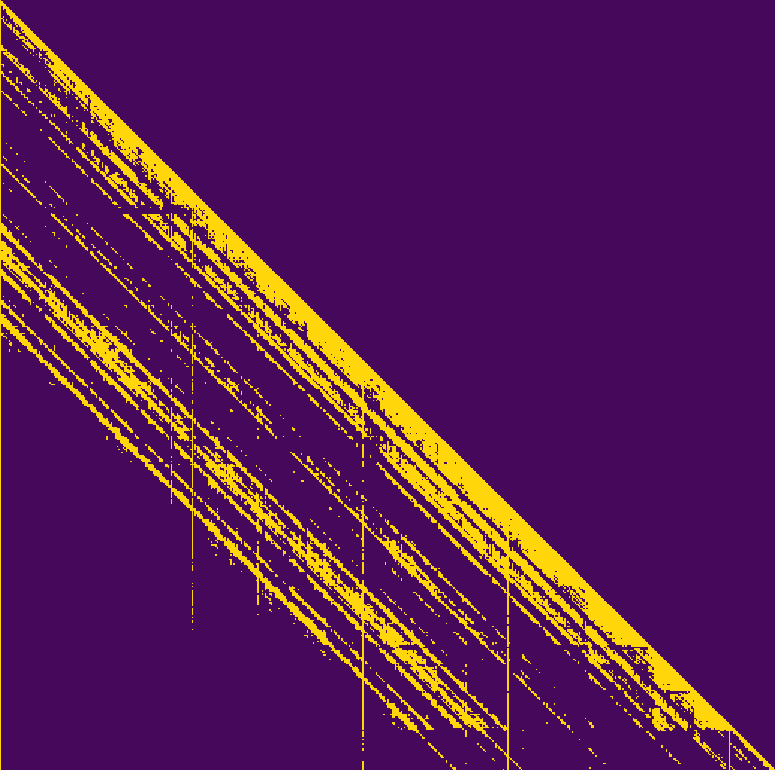}
    }
    \caption{Attention pattern visualization at 64K context length. Each pair shows FullAttention ground truth (left) and RRAttention selection (right).}
    \label{fig:vis_64k}
\end{figure*}
\begin{figure*}[htp]
    \centering
    \subfloat[FullAttention]{
        \includegraphics[width=0.24\textwidth]{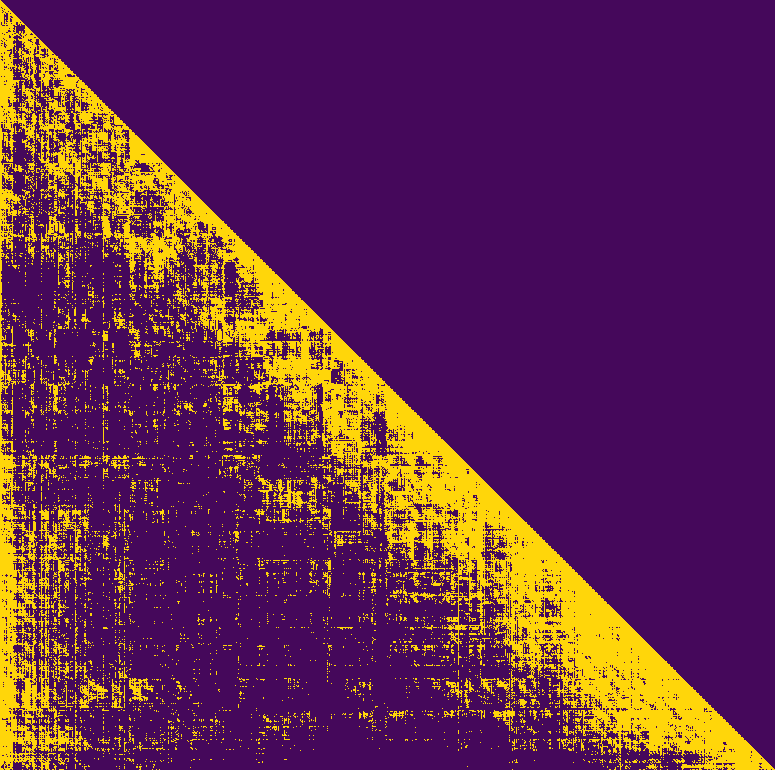}
    }%
    \subfloat[RRAttention]{
        \includegraphics[width=0.24\textwidth]{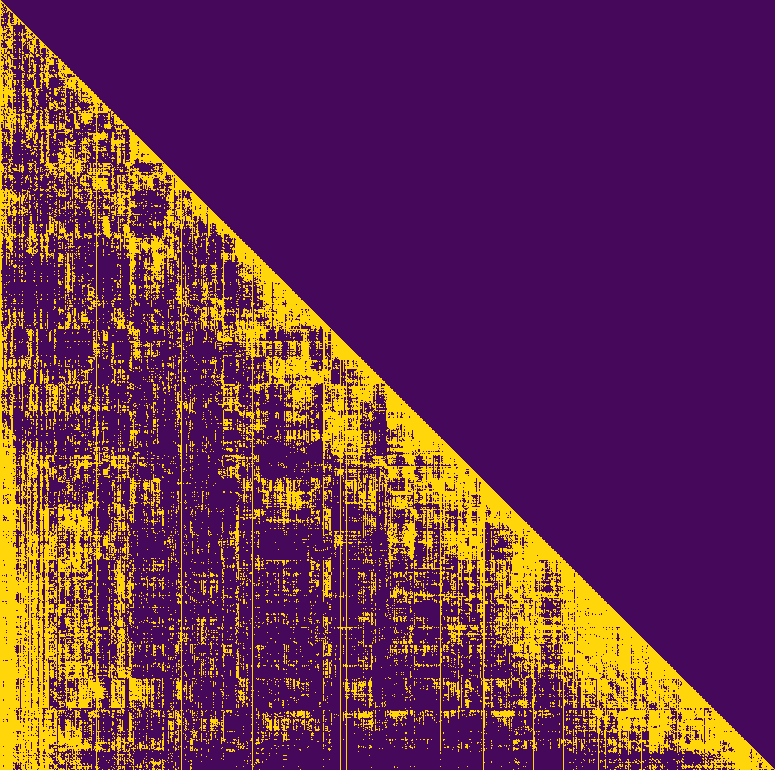}
    }%
    \subfloat[FullAttention]{
        \includegraphics[width=0.24\textwidth]{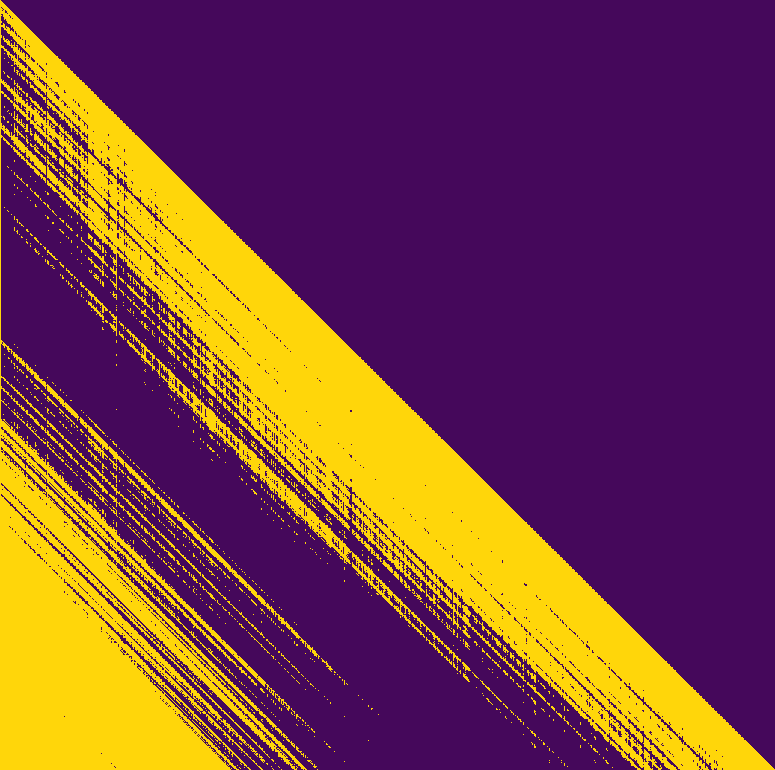}
    }%
    \subfloat[RRAttention]{
        \includegraphics[width=0.24\textwidth]{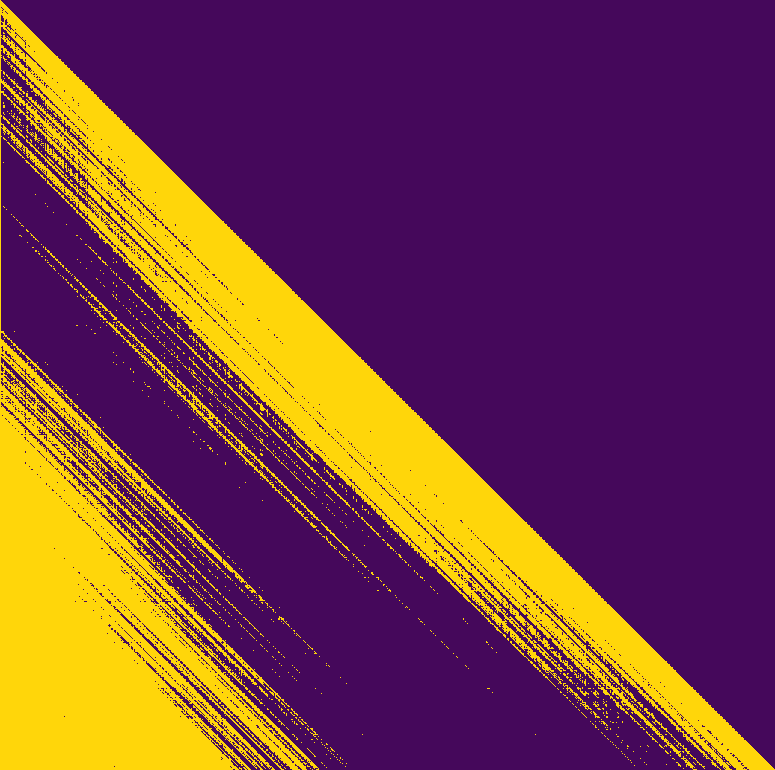}
    }
    \caption{Attention pattern visualization at 128K context length. Each pair shows FullAttention ground truth (left) and RRAttention selection (right).}
    \label{fig:vis_128k}
\end{figure*}

As shown in Figures~\ref{fig:vis_16k}-\ref{fig:vis_128k}, RRAttention effectively captures diverse attention patterns across different context lengths. The visualizations reveal several typical patterns: 
(1) \textit{local patterns} concentrated around the main diagonal, (2) \textit{vertical and slash patterns} focusing on specific important tokens (e.g., special tokens or semantically important positions) and diagonal structures, and (3) \textit{scatter patterns} manifesting as distributed points across the sequence, representing global information retrieval.
Across all context lengths, RRAttention's selected blocks (right column of each pair) closely match the ground truth important blocks derived from FullAttention (left column of each pair), demonstrating that our stride-level importance estimation and block-level selection accurately identify the critical attention regions. The high visual similarity between FullAttention ground truth and RRAttention selection validates our method's ability to preserve essential attention patterns while achieving significant sparsity.

\section{Future Work}
\label{sec:future_work}
While RRAttention demonstrates strong performance in long-context scenarios, several promising directions remain for future work to further enhance its efficiency and applicability.

\textbf{Engineering Optimization.} Our current implementation is based on FlashAttention-2 sparse kernels. Migrating to FlashAttention-3 would leverage its advanced features, including warp specialization, improved memory access patterns, and better GPU occupancy. These architectural improvements could further reduce latency and increase throughput without any algorithmic changes, representing a direct path to enhanced performance across different hardware platforms.

\textbf{Training-Aware Sparse Attention.} Currently, RRAttention performs dynamic pattern discovery at inference time, which incurs runtime overhead for pattern search. An alternative paradigm is to incorporate sparse attention learning during the training phase, enabling models to internalize optimal sparsity patterns through distillation from full attention or learnable gating mechanisms. This training-aware approach could eliminate pattern discovery overhead entirely while potentially achieving higher sparsity levels with maintained accuracy, shifting from runtime discovery to learned prediction.

\textbf{Extension to Decoding Stage.} RRAttention currently focuses on the prefill stage, where the head round-robin sampling strategy effectively identifies important attention patterns with significant sparsity. This principle can be naturally extended to the decoding stage to reduce KV cache memory bandwidth consumption and lower per-token latency. However, the sequential nature of decoding presents unique challenges, requiring careful design to preserve generation quality while maintaining query independence.
These directions collectively aim to make long-context inference more efficient across the entire generation pipeline, from initial context processing to final token generation.

\end{document}